\documentclass[journal]{IEEEtran}
\usepackage[utf8]{inputenc}
\usepackage[fleqn]{amsmath}
\usepackage{graphicx}
\usepackage[ruled]{algorithm}
\usepackage[noend]{algpseudocode}
\usepackage{latexsym}
\usepackage{cite}

\ifCLASSINFOpdf
  \DeclareGraphicsExtensions{.pdf,.jpeg,.png}
\else
\fi
\ifCLASSOPTIONcompsoc
 \usepackage[caption=false,font=normalsize,labelfont=sf,textfont=sf]{subfig}
\else
  \usepackage[caption=false,font=footnotesize]{subfig}
\fi
\begin{document}
%
\title{FALCON: Feature Driven Selective Classification for Energy-Efficient Image Recognition}
%
%
%

\author{Priyadarshini~Panda, \textit{ Student Member}, \textit{IEEE}, Aayush~Ankit, Parami~Wijesinghe,
        and Kaushik~Roy,~\IEEEmembership{Fellow,~IEEE}
\thanks{P. Panda, A. Ankit, P. Wijesinghe and K. Roy are with the School
of Electrical and Computer Engineering, Purdue University, West Lafayette,
IN, 47907 USA.
Correspondance E-mail: pandap@purdue.edu.}
}

%
%

\markboth{Accepted for publication in IEEE Transacations on Computer-Aided Design of Integrated Circuits and Systems,~2017}%
{Shell \MakeLowercase{\textit{et al.}}: Bare Demo of IEEEtran.cls for IEEE Journals}
%



\maketitle

\begin{abstract}
Machine-learning algorithms have shown outstanding image recognition/classification performance for computer vision applications. However, the compute and energy requirement for implementing such classifier models for large-scale problems is quite high. In this paper, we propose \underline{F}e\underline{a}ture Driven Se\underline{l}ective \underline{C}lassificati\underline{on} (FALCON) inspired by the biological visual attention mechanism in the brain to optimize the energy-efficiency of machine-learning classifiers. We use the consensus in the characteristic features (color/texture) across images in a dataset to decompose the original classification problem and construct a tree of classifiers (nodes) with a generic-to-specific transition in the classification hierarchy. The initial nodes of the tree separate the instances based on feature information and selectively enable the latter nodes to perform object specific classification. The proposed methodology allows selective activation of only those branches and nodes of the classification tree that are relevant to the input while keeping the remaining nodes idle. Additionally, we propose a programmable and scalable Neuromorphic Engine (NeuE) that utilizes arrays of specialized neural computational elements to execute the FALCON based classifier models for diverse datasets. The structure of FALCON facilitates the reuse of nodes while scaling up from small classification problems to larger ones thus allowing us to construct classifier implementations that are significantly more efficient. We evaluate our approach for a 12-object classification task on the Caltech101 dataset and 10-object task on CIFAR-10 dataset by constructing FALCON models on the NeuE platform in 45nm technology. Our results demonstrate up to 3.66x improvement in energy-efficiency for no loss in output quality, and even higher improvements of up to 5.91x with 3.9\% accuracy loss compared to an optimised baseline network. In addition, FALCON shows an improvement in training time of up to 1.96x as compared to the traditional classification approach.
\end{abstract}

\begin{IEEEkeywords}
Machine Learning, Hierarchical Feature Learning, Energy-Efficient Classification, Selective Activation, Neuromorphic Hardware.
\end{IEEEkeywords}
%

\section{Introduction}
Machine-learning classifiers have proven to be very successful for several cognitive applications such as search, classification, recognition \cite{dubey2005recognition,jones2014learning,netzer2011reading} among others and are being increasingly deployed across a wide range of computing platforms from data centers to mobile devices. While the classifiers are modeled to mimic brain-like cognitive abilities, they lack the remarkable energy-efficient processing capability of the brain. For instance, SuperVision \cite{krizhevsky2012imagenet}, a state of the art deep learning Neural Network (NN) for image classification tasks, demands compute energy in the order of 2-4 Giga-OPS (Multiply and Accumulate operations (OPS)) per classification \cite{ramasubramanian2014spindle}, which is nearly 8$\sim$9 orders of magnitude larger than the human brain. With energy efficiency becoming a primary concern across the computing spectrum, energy-efficient realization of large-scale neural networks is of great importance. 

It is well known that the visual cortical system is arranged in a hierarchical fashion with different areas responsible for processing different features (for example, color and shape) of visual information \cite{whitney2009neuroscience,ungerleider2000mechanisms}. For a given input, the visual information is decomposed into representative features and only those areas of the brain that are instrumental to the recognition of the input are activated. The innate ability to simplify complex visual tasks into characteristic features and the selective activation of different areas based on the feature information in the input, enables the brain to perform cognition with extremely low power consumption. In this paper, we build upon this biological concept of feature selective processing to introduce Feature driven Selective Classification (FALCON) for faster and energy-efficient image recognition with competitive classification accuracy. 

Interestingly, we note that there is a significant consensus among features of images across multiple classes in a real world dataset. Consider the simple classification problem of recognizing 4 different objects: strawberry, sunflower, tennis ball and stop sign. All 4 objects belong to completely different classes. However, strawberry and stop sign have a feature i.e. the red color as representative information common across all images of the 2 objects. Similarly, sunflower and tennis ball have the characteristic yellow color as a common feature. Here, we utilize the feature consensus to break up the classification problem and use a cluster of classifiers to perform smaller classification tasks. We achieve this by constructing a hierarchical tree of classifiers wherein the initial nodes (or classifiers) are trained first to classify the image into general feature categories: red and yellow (for the above example), while the deeper nodes categorize them into the 4 specific classes. The generic-to-specific transition in the classification hierarchy enable us to selectively process only those branches and nodes that are relevant to the input. 

\begin{figure}[t!]
\centering
\includegraphics[width=0.5\textwidth]{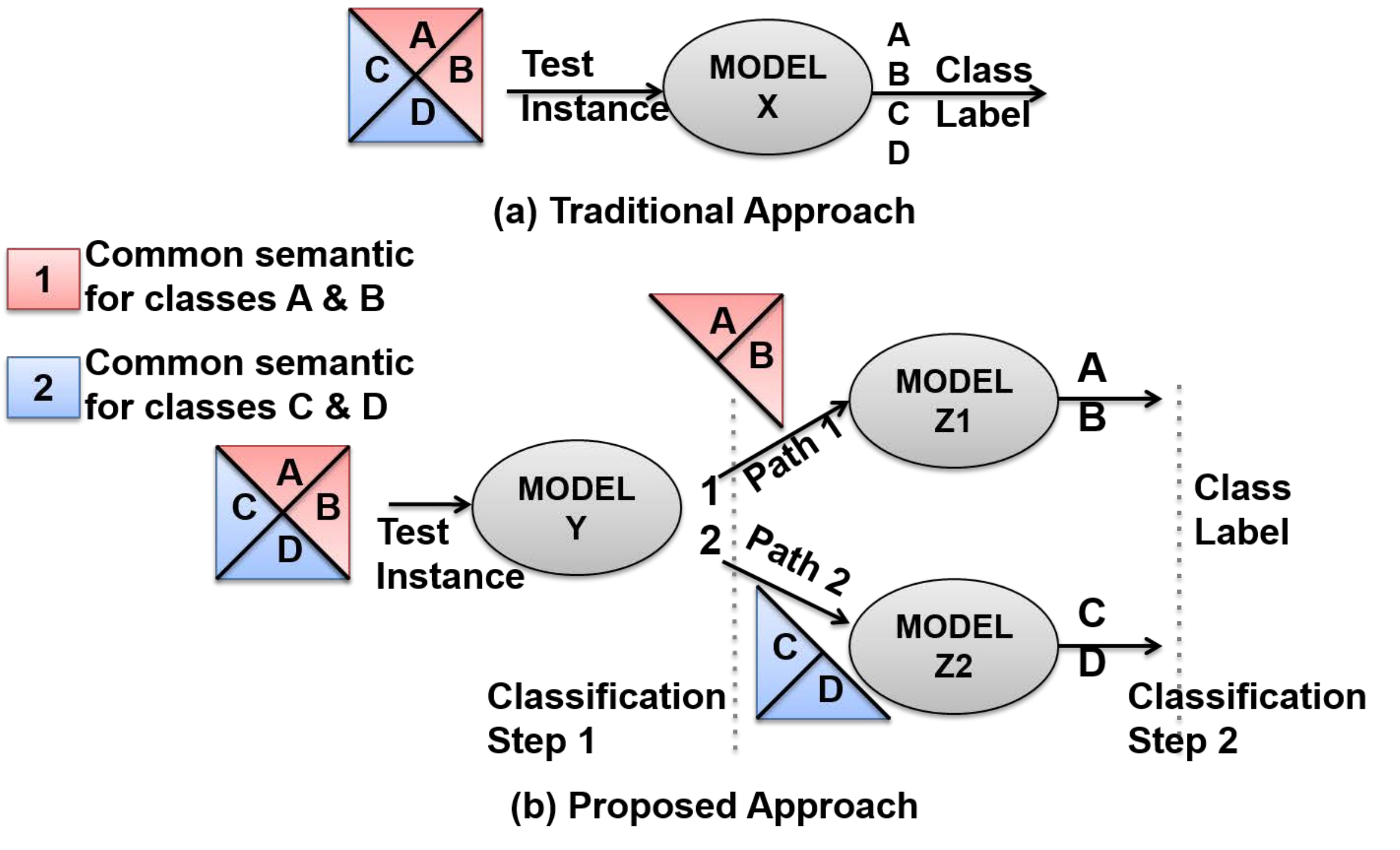}
\caption{(a) Traditional approach where a single model is applied for classifying instances into 4 classes (b) Proposed FALCON approach where 4-object classification problem is decomposed into simpler tasks based on feature consensus between inputs. The paths corresponding to the specific feature are selectively activated.}
\end{figure}

Fig. 1 illustrates our methodology. In the traditional approach shown in Fig. 1(a), a single classifier is responsible for classifying the inputs into the 4 distinct classes (A, B, C, D). Hence, the network clearly needs to be highly complex (with more neurons and synapses) in order to classify the objects with high accuracy. However, this Model X does not take into account the common features across classes and thus expends constant computational effort on all inputs activating each and every connection/neuron to determine the output. In contrast, Fig. 1(b) shows our proposed FALCON approach wherein we build a hierarchical tree of classifiers based on the feature consensus between classes (A, B and C, D). The initial node (Model Y) in the tree is trained to distinguish between the features (1 \& 2). The latter nodes (Model Z1, Z2) perform the final classification task of separating the objects into classes A, B (Model Z1) and C, D (Model Z2). Since these models (Y, Z1, Z2) are trained to classify between two different classes, they will be less complex than the traditional Model X. It can be clearly seen that the classification task is now broken down into a 2-step process which involves two different paths comprising of separate nodes. Due to the 2-step classification, Z1 and Z2 need to be trained only on a subset of the training dataset as shown in Fig 1(b), resulting in significant reduction in the training time of these nodes. For a given input instance, if Model Y gives a high confidence at output neuron P (Q), then, only path 1 (2) and the corresponding Model Z1 (Z2) is enabled while keeping Z2 (Z1) idle. Hence, our approach is both time and energy efficient, since it involves selective activation of nodes depending upon the input instance. 

Another significant contribution of our work is the design of a scalable Neuromorphic Engine (NeuE) that provides a programmable hardware platform for executing FALCON models with various nodes and weights. The neuromorphic engine features a 1D array of Neural Units (NUs) followed by an Activation Unit (AU) that process the basic computational elements of neural networks. We enable the NeuE with appropriate hardware mechanisms to effectively implement selective activation of nodes for energy benefits at run-time. 

In summary, the key contributions of this work are as follows:
\begin {itemize}
\item Given any machine learning classifier, we propose a systematic methodology to construct a feature driven selective classification framework that exploits the consensus in the characteristic features (color/texture) across images in a dataset to perform faster and energy-efficient classification. The methodology is independent of the network topology, network parameters and training dataset. 
\item We develop a design methodology to construct a tree of classifiers (or nodes) with a generic-to-specific transition in the classification hierarchy invoking multi-step classification. The initial nodes of the tree separate the instances based on feature information and selectively enable the latter nodes to perform object specific classification. 
\item In this work, we use color and texture as the distinctive features to implement FALCON. We also present an algorithm to select the optimal color/textures common across multiple classes of objects. \item We design a programmable and scalable Neuromorphic Engine (NeuE) that can be used to efficiently execute FALCON models on Artificial Neural Networks (ANNs).
\item We demonstrate the efficacy of our proposed approach on two natural image datasets: Caltech101/ CIFAR10. We construct the FALCON based hierarchical tree of ANNs using the proposed design methodology and execute them on the NeuE platform to demonstrate significant improvements in energy for negligible loss in output quality.
\end{itemize}

The rest of the paper is organized as follows. In Section II, we discuss related work. In Section III, we present the structured approach to construct FALCON models. Section IV details the architecture of NeuE. Section V describes the experimental methodology and the benchmarks. We discuss the results in Section VI and conclude in Section VII. 

\section{Related Work}
The widespread use of machine learning across computing platforms from data centers to mobile devices has renewed interest in forming efficient methodologies for classification that expend low compute effort. On the algorithmic front, substantial work for increasing accuracy in machine-learning classification has been done \cite{deng2014ensemble,sun2013deep}. Using semantics or feature information for improving the accuracy of content based image retreival systems has been an active area of research \cite{liu2007survey}. In \cite{smeulders2000content}, a comprehensive review of various techniques geared towards extracting global image features (color, texture, local geometry) for accurate image retreival has been discussed. The key idea of using high-level semantic features in our proposed FALCON methodology is inspired from content based systems. However, the novelty of our work arises from the fact that we leverage the similarity in the features across various classes for clustering several classes into one and thus decomposing a large classifcation problem into smaller tasks organised in a tree-fashion to obtain efficiency in training as well as testing complexity. 

Recently many decision tree methods for large scale classification have been proposed. The first group of methods do not assume that classes are organized into a hierarchy. It includes methods based on “one-versus-all” and “one-versus-one” strategies, which further assume classes are unrelated (e.g., do not share features). It also includes error correcting output codes \cite{allwein2000reducing, torralba2004sharing}, which utilize the relationship between classes (e.g., sharing features) to build more compact and robust models. These methods typically show good classification accuracy. However, the time complexity for evaluating the classifiers are “linearly” proportional to the number of classes. 

The second group of methods aims at reducing the time complexity utilizing the hierarchical structure of classes. In \cite{bengio2010label, deng2011fast, beygelzimer2009conditional, rastegari2012attribute}, the authors propose different methods to automatically build the hierarchy of classes. Other methods \cite{zweig2007exploiting} rely on a given hierarchy. However, in order to achieve fast evaluation, such tree-based methods exploit the hierarchical structure in the label space by organizing a set of binary classifiers where each binary classifier consists of two subsets of classes. The binary partition of classes at each node does not lead to good separability, especially for the difficult instances or classes in the dataset at the initial nodes, causing a decline in accuracy. 

While FALCON is related to such tree-based methods, the fundamental feature selection methodology to cluster groups of classes does not restrict the partitioning of classes into two primary groups. As a result, the decision boundary model created at the initial nodes of the tree is more flexible that can handle difficult classes in the dataset accurately. In conventional tree-based methods, each tree hierarchy constructed is very specific to the given dataset. Thus, for every new class or object that has to be added to the classification problem, each node of the tree has to be retrained with the additional classes, which significantly increases the training cost. In contrast, the structure of FALCON enables us to reuse nodes while scaling up from small classification problems to larger ones, thereby reducing the training complexity and also making the methodology scalable for hardware implementations. 

In the recent past, there has been significant work employing approximate computing techniques to obtain efficient neural computations relying on the error resilient properties of recognition applications \cite{panda2016invited}. In \cite{venkataramani2014axnn}, the authors have considered domain specific insights to introduce hardware approximations in neuromorphic applications. In \cite{panda2015object, panda2017energy}, the authors have utilized the inherent feature variability across input instances in a dataset to introduce software techniques for designing energy-efficient scalable classification framework. In the context of efficient neuromorphic systems, two major directions have been explored. The first is accelerator based computing where custom architectures for specific computation of NNs are designed. In \cite{chakradhar2010dynamically,chen201614}, application-specific NN designs and programmable neuromorphic processors have been proposed. Also, NN implementations on programmable accelerators such as GPUs have also been explored \cite{ngiam2011optimization}. The second is the use of emerging post-CMOS device such as resistive RAM \cite{rajendran2013specifications}, memristive crossbars \cite{jo2010nanoscale} and spintronics \cite{roy2013beyond}, to realize the individual computational elements: neurons and synapses more efficiently.

In this work, we propose a new avenue for energy efficiency in neuromorphic systems by using representative features across images in a real-world dataset. The main focus of this paper is in developing an automatic design methodology to generate FALCON models to lower the testing complexity in traditional classification problems. In contrast to the approximate techniques \cite{panda2016invited} that usually provide an explicit tradeoff between efficiency and quality of results, our approach maintains classification accuracy while providing energy savings. In addition, our design methodology provides the opportunity to reuse nodes (discussed in Section III) enabling the classification framework to be more scalable. Note that the efforts on efficient neuromorphic systems mentioned earlier can be employed with our proposed design methodology to further enhance the efficiency. Also, our methodology improves the training time for large classification tasks which is one of the major challenges in machine learning at present.

\section{Feature Driven Selective Classification (FALCON): Approach and Design}
In this section, we present our structured approach to construct FALCON based hierarchical tree of classifiers. While there exists a suite of machine-learning classifiers like Support Vector Machines, Decision trees, Neural Networks etc. suitable for classification, we will focus on a particular class: Artificial Neural Network (ANNs) to validate the proposed methodology for image recognition. Please note that the FALCON tree can be applied on other machine-learning algorithms as well to lower the compute energy. 

\subsection{Feature Selection from Input}
FALCON employs the features, representative of the input image data, to construct the nodes of the hierarchical tree. Referring to Fig. 1, Model Y is trained to classify the inputs based on the feature information. Hence, the appropriate selection of features is crucial. While there can be several image features that can be used to discriminate the first step of selective  classification, in this work, we use color and texture as our distinctive features to implement FALCON. In fact, texture and color are the most widely usedrepresentative features for characterizing low-level image information \cite{singha2012content}. In this work, we use Hue-Saturation-Value (HSV) transformation \cite{levkowitz1993glhs} and Gabor filtering \cite{jain1997object} to extract the color and texture features of an image, respectively. Applying HSV or Gabor filtering onto an image results in dimensionality reduction of the original image. The reduced feature vector contains the relevant feature information, which is sufficient to characterize and classify an image. Traditionally, images are transformed with appropriate feature extraction techniques to get a lower dimensional input vector \cite{palm2000gabor}. A machine-learning classifier yields better classification accuracy and converges to global minima faster when trained on the feature vector as opposed to the original input image. Since FALCON invokes multi-step classification, it therefore, enables the latter nodes in the tree (Model Z1, Z2 in Fig. 1) to be trained on feature vectors alone, instead of real pixel valued images. Due to the significant reduction in the input vector size, the models Z1 and Z2 are much simpler (fewer neurons and connections) as compared to the traditional model X. Please note that we need to take into account the additional computational cost of HSV and Gabor filtering for calculating energy costs \cite{chen2008fast,bernardino2005real}.  

HSV gives rise to feature vectors corresponding to 8 color components per image. Similarly, Gabor filters corresponding to ‘m’ scales and ‘n’ orientations give rise to m x n texture components per image \cite{haghighat2013identification}. In this work, we use Gabor filters with scales: 4$\sqrt{2}$ * i \{i= 1,2,4,8\} and 4 orientations: 0, 45, 90, 135 degrees, which are adequate for characterizing the basic texture components of an image \cite{jain1997object}. For each orientation, the texture features across all scales (4$\sqrt{2}$ * i \{i= 1,2,4,8\}) are concatenated into a single feature vector. So, the feature selection methodology identifies the most probable orientation across the set of concatenated texture vectors. The most important question that needs to be answered is how we select the optimal features (color/texture) to categorize the images in a dataset into the general feature classes. We employ a simple search-based method to obtain the features common across multiple classes of objects. 

\begin{figure}[t!]
\centering
\includegraphics[width=0.5\textwidth]{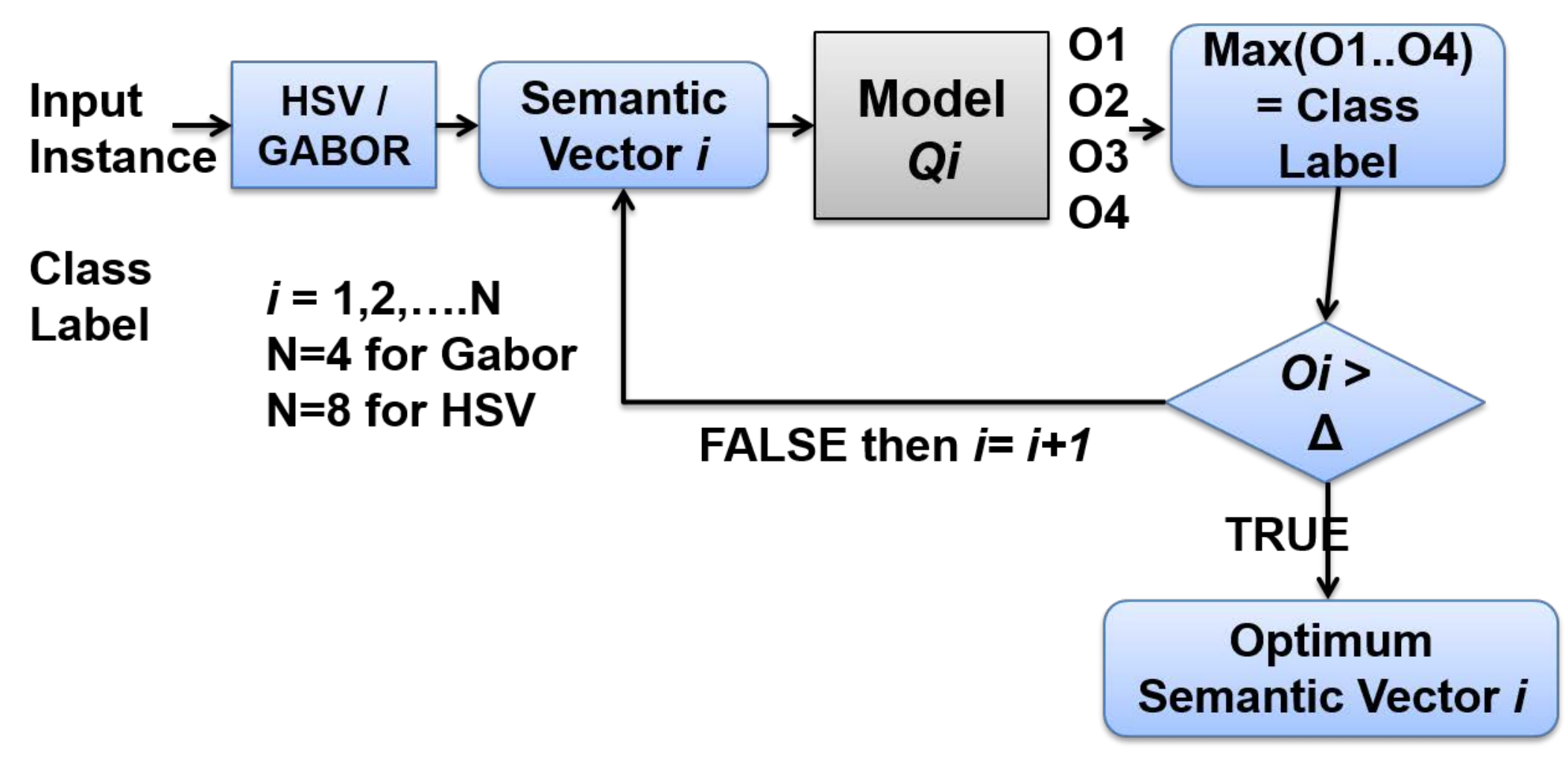}
\caption{An overview of the Feature Selection methodology for a dataset with 4 classes.}
\end{figure}
Fig. 2 gives an overview of the feature selection methodology for a dataset with 4 distinct classes. For each class of objects in a dataset, we train a NN (Model $Q_i$) based on a particular feature (feature vector i) with the target labels provided with the dataset. This is done for all four texture (corresponding to the 4 orientations with scales concatenated) and the eight color components. In each case, the NN’s size and the number of iterations remain fixed. Once the models corresponding to each feature are trained, we pass a single input image for a given class through each model. The feature that gives the highest confidence value ($O_i$) at the output is chosen as the optimum one for that particular class, given that the confidence value is above a certain user-defined threshold $\Delta$. For instance, in the sunflower/strawberry/tennis/ stop-sign classification problem, applying the above method across all 4 classes we obtain that Red feature produces a confidence value of 0.9 and 0.8 for Strawberry and Stop-Sign while 0.3 and 0.2 for tennis and sunflower, respectively. Thus, Strawberry and Stop-Sign will be categorized under the Red category by the initial node (Model Y from Fig. 1). $\Delta$ is chosen to be around 0.6-0.8 to get the most accurate feature selection. 

\subsection{FALCON: Preliminaries}
\subsubsection{Structure of the FALCON tree}
\begin{figure}[t!]
\centering
\includegraphics[width=0.5\textwidth]{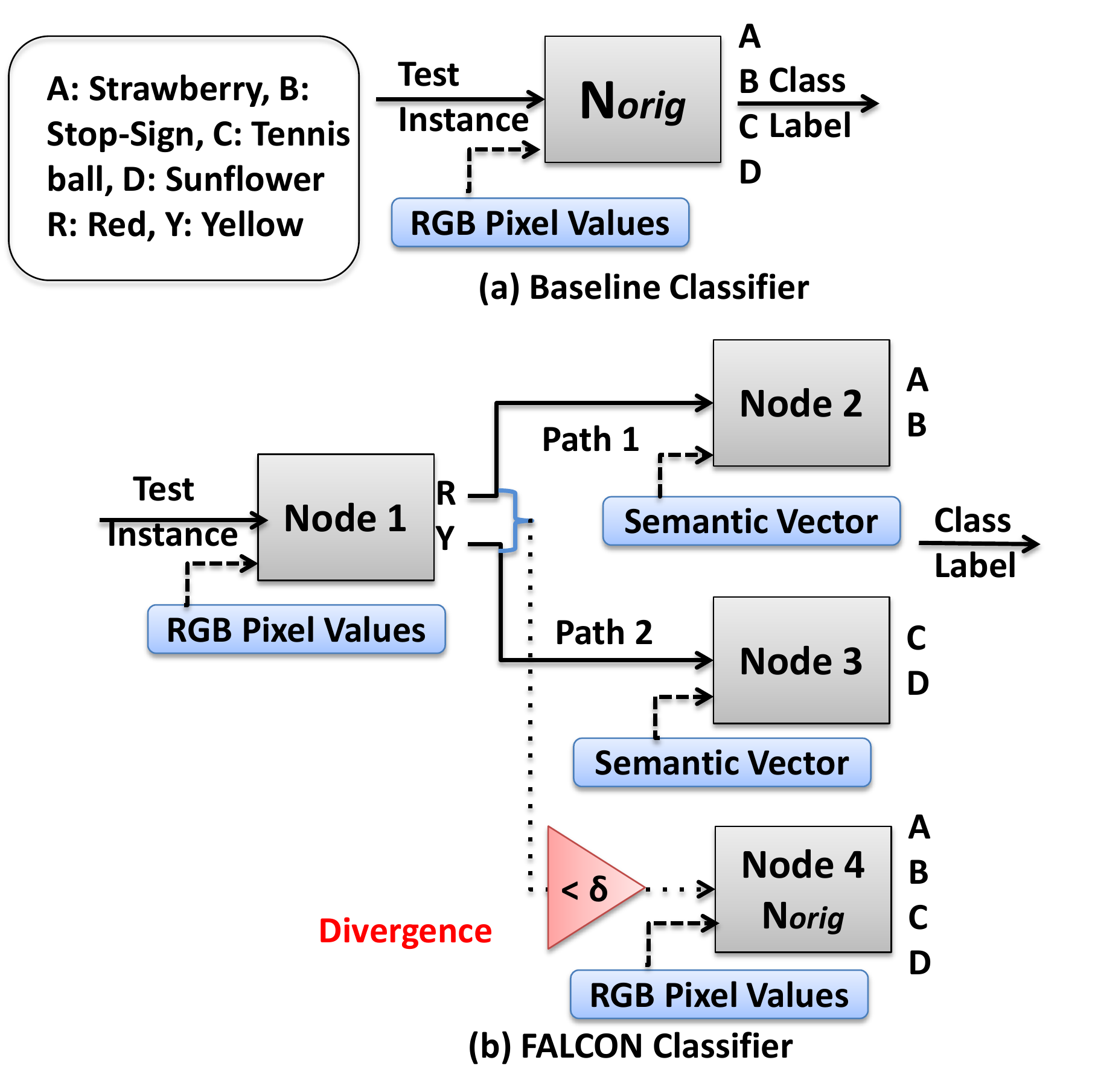}
\caption{(a) Baseline single NN Classifier (b) FALCON classifier with 4 nodes where the output of the initial node (Node 1) is monitored to selectively enable Node 2, 3 and 4.}
\end{figure}
Fig. 3 shows the conceptual view of the framework for a 4-object classification problem. Fig. 3(a) shows the baseline classifier with a single NN that has 4 output neurons corresponding to each class (A, B, C, D). Fig. 3 (b) illustrates the proposed FALCON based tree with three nodes (not considering Node 4 for now). Each node is a NN classifier trained using the standard backpropagation algorithm. First, the feature selection methodology discussed in Section III(A) is employed to obtain the general features that are used as training labels (R, Y) for the initial node (Node 1). Node 1 is responsible for classifying the input into the two broad feature categories and thus has two output neurons. Node 2 and 3 then separate the inputs with feature consensus into the corresponding classes (A, B and C, D).  Thus, the class labels produced at Node 2 and 3 are expressed as the final output of the FALCON framework. Node 2 (3) is selectively activated only if the class label produced from Node 1 is R (Y). Node 2 and 3 are trained on the reduced feature vectors as input. In contrast, the original RGB pixel values are fed as input to Node 1 to obtain a competitive classification accuracy with respect to the baseline classifier. The multi-step classification process enables the nodes in the FALCON tree to be less complex than the baseline NN resulting in overall energy-efficiency. 

\subsubsection{Accuracy Optimization}
 In FALCON, each node of the tree is trained separately on the input instances as discussed above. During test time, data is processed through each node in the tree to produce a class label. It is evident that the initial node (Node 1 in Fig. 3(b)) of the FALCON tree would be the main bottleneck for achieving iso-accuracy with that of the baseline classifier. For an input instance belonging to Class R, if Node 1 produces a higher confidence value for Class Y, the input instance is not passed to the latter nodes and is misclassified at the first stage itself, resulting in a decline in accuracy. This would arise when the input instance has characteristics pertaining to both features (R and Y). For example, an image of a strawberry might have some yellow objects in the background. In such cases, the difference in the confidence of the two output neurons at Node 1 would be low. As a result, the instance will get misclassified. To avoid this, we add the baseline classifier as a 4th node in the FALCON tree that is enabled by the divergence module (triangle in Fig. 3(b)). The divergence module activates the 4th node if the confidence difference at the outputs of initial node is below a certain \textit{divergence value}, $\delta$. In that case, the paths 1 and 2 of the tree are disabled. This is in accordance with the selective processing concept. Later, in Section V(A), it is shown that accuracy degradation with respect to baseline in the absence of the divergence module (or the baseline node) in the FALCON tree is around 2-4\% for most classification problems. Thus, for applications where the slight degradation in accuracy is permissible, it is not required to append the baseline classifier to the FALCON tree. 

\subsubsection{Node Reusability}

\begin{figure}[t!]
\centering
\includegraphics[width=0.5\textwidth]{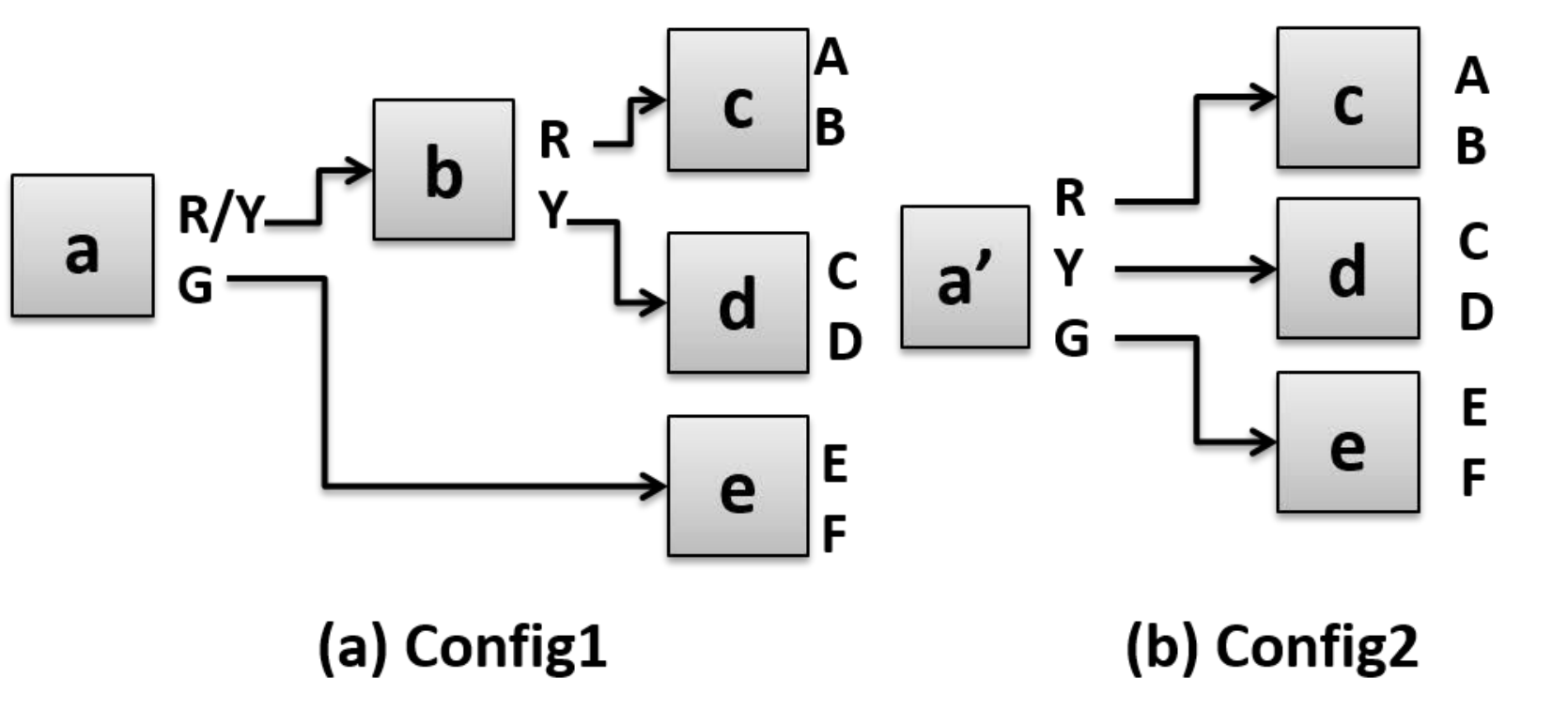}
\caption{Two different configurations of FALCON tree for a 6-object classification problem reusing Nodes (2/3=c/d) from Fig. 3(b). }
\end{figure}
FALCON facilitates the reuse of nodes (or classifiers) from one classification tree to another when we want to incorporate additional classes with new feature information into a given task. Consider a 6-object classification problem wherein 4 classes are the same as that of Fig. 3(b) and the remaining two new classes (E, F) have a common feature G. There are two different ways of constructing the FALCON tree for the given problem as shown in Fig. 4. We have not shown the divergence module for the sake of convenience in representation. It is evident that the last nodes (c, d, e) which provide the final output of the classifier are the same in both Fig. 4(a), (b). Additionally, the nodes c, d, b are the same as that of nodes 1, 2, 3 (Fig. 3(b)), respectively. Hence, these nodes (1, 2, 3) from Fig. 3(b) can be reused for the 6-object problem where learning the weights for these nodes is not required. FALCON allows us to create reusable models (trained for a particular classification problem) and use the same for different classification problems. Reusability is one of the major benefits that FALCON provides over conventional algorithms. In the conventional approach, the NN has to be retrained whenever a new class or object is added to the classification problem. For instance, the baseline NN in Fig. 3(a) needs to incorporate 6 neurons at the output layer in this case. As the networks are fully connected, the weights have to be learnt all over again to achieve a nominal accuracy. In a resource-constrained environment, reusability with FALCON would enable us to realize large-scale classification frameworks in hardware, addressing more challenging problems in an energy-efficient manner. Node reusability thus provides the FALCON methodology with the added advantage of scalability. 
\setlength{\textfloatsep}{0pt}
\begin{algorithm}[t]
\caption{Methodology to construct FALCON tree}
\label{algo1}
 \textbf{Input:} Training dataset with the target labels ($l_t$) for each class ($t$), Baseline classifier ($N_{orig}$)\\
 \textbf{Output:} FALCON Tree ($N_{ssc}$)
\begin{algorithmic}[1]
\State Obtain the relevant features associated with each class/object ($t$) in the dataset with the feature selection methodology described in Section III (A).  
\State Group the objects and the corresponding training labels ($l_t$) with feature consensus under one label ($l_i$). The labels ($l_i$) serve as training labels for the initial node. 
\newline\textbf{initialize} count= \# of labels ($l_i$) obtained, $output_i$= \# of classes ($t$) grouped under $l_i$ 
\State Train the initial node ($node_i$) of the FALCON tree based on the labels ($l_i$) to classify the objects based on their features. \# of output neurons in $node_i$= count. 
\newline \textit{The input vector at $node_i$ is the original RGB pixel values of the image.}
\State \textbf{initialize} \# of final nodes ($node_f$) in the tree = $count$. 
\State \textbf{for} $ j =1:count$ // \textit{for each node based on the feature concensus}
\State Train $node_f(j)$ with target labels ($l_t$) corresponding to classes with feature consensus.
\newline  \# of output neurons in $node_f(j)$ = $output_i(j)$. \textit{The input vector at $node_f(j)$ is the feature vector of the image.}
\State \textbf{end for}
\State Append $N_{orig}$ as the last node to $N_{ssc}$ depending upon the accuracy requirement. Please note that each node of the FALCON tree is trained to achieve iso-accuracy with that of the baseline.
\newline \textit{Please note that each node of the FALCON tree is trained to achieve iso-accuracy with that of the baseline.}
\end{algorithmic}
\end{algorithm}
\subsubsection{Energy Optimization}
There are different ways of constructing a FALCON tree for a given classification task. However, we need to select the configuration that yields higher energy savings without compromising the output accuracy significantly. Referring to the 6-object classification problem described above, both configurations in Fig. 4 will yield computational savings with respect to the baseline NN as it invokes selective activation of various nodes in the tree. However, the configuration in Fig. 4(a) (\textit{Config1}) would yield higher energy savings than that of Fig. 4(b) (\textit{Config2}) for a drastic accuracy degradation. This can be explained as follows: It is evident that Node \textit{a’} in \textit{Config2} will be slightly larger than Node\textit{a} in \textit{Config1} as there are more number of output classes to identify in the former case. However, when we merge instances that can be identified with two different features into one (similar to \textit{R/Y} in \textit{Config1}), then, almost $2/3^{rd}$ of the dataset (all instances belonging to Class \textit{A,B,C,D}) is being classified into one category (i.e. \textit{R/Y}) at Node \textit{a}. This gives rise to an imbalanced dataset for the first node (Node \textit{a}). In our experiments, we saw that due to this imbalance, the NN at Node \textit{a} was often biased towards the majority class (i.e \textit{R/Y} in this case) that resulted in a higher error rate for the minority class (\textit{G} in Node \textit{a}). Specifically, instances from classes \textit{E, F} that should ideally be classified as Class \textit{G} at Node \textit{a} were falsely classified as Class \textit{R/Y} causing a decline in overall accuracy. Also, from the efficiency perspective, these classes that could have been identified with two nodes (Node \textit{a, e}) wrongly activate 3 nodes (Nodes \textit{a, b, c/d}) leading to higher computational costs. As mentioned earlier in Section II with regard to decision tree based classification, the binary partition of classes at the node does not lead to good separability that causes an accuracy decline.

In contrast, in \textit{Config2}, although Node \textit{a’} is slightly more computationally expensive than Node \textit{a/b} in \textit{Config1}, the fact that we do not restrict the partition of classes into two partitions leads to a good separability as the dataset is now balanced for Node \textit{a’}. Consequently, the instances of each category are identified correctly at the initial node and the corresponding path to the specific classifier at latter nodes is activated. Thus, \textit{Config2} does not degrade the accuracy. Also, the instances from classes \textit{E, F} in \textit{Config2} require activation of only two nodes (Node \textit{a’, e}). The energy expended for the misclassified instances at \textit{Config1} (false activation of Nodes \textit{a, b, c/d}) exceeds the slightly higher computational cost imposed by Node \textit{a’} in \textit{Config2}. Thus, \textit{Config2} serves as a more energy-efficient option than \textit{Config1} that doesn’t degrade the accuracy as compared to the baseline. \textit{Config1} will be energy-efficient with severe loss in accuracy that is generally not permissible in real-time applications. Thus, for a given classification problem, FALCON tree with initial node for feature classification and final nodes for object-specific classification (as in \textit{Config2}) would yield maximum benefits without conceding the classification accuracy.


\subsection{FALCON:Design and Testing Methodology}
The systematic methodology to construct the FALCON tree is given in Algorithm 1. The process takes a pre-trained baseline classifier (single NN, $N_{orig}$), its corresponding training dataset with the target labels ($l_t$) as input, and produces a FALCON tree ($N_{ssc}$) as output. 

Once the FALCON tree is constructed, we input the test data to the tree to obtain accuracy and efficiency results. The overall testing methodology is shown in Algorithm 2. Given a test instance $I_{test}$, the process obtains the class label $L_{test}$ for it using the FALCON tree ($N_{ssc}$). The output from the initial node is monitored by the divergence module to decide if a path of the tree corresponding to a final node ($node_f$) or the baseline classifier $N_{orig}$) is to be activated. 

\begin{algorithm}[t]
\caption{Methodology to test FALCON tree}
\label{algo1}
 \textbf{Input:}Test Instance $I_{test}$, FALCON Tree ($N_{ssc}$)
 \textbf{Output:} Class Label. $L_{test}$
\begin{algorithmic}[1]
\State Obtain the feature vectors for $ I_{test}$  corresponding to the labels ($l_i$) obtained for the initial node ($node_i$). 
\State Obtain the output of $node_i$  and compute the difference between the maximum ($o_{max}$) and minimum ($o_{min}$) confidence values across all output neurons of $node_i$. 
\State \textbf{if} $|o_{max}$ - $o_{min} | < \delta$ (user-defined divergence value) \textbf{then} enable baseline classifier ($N_{orig}$). Class Label $L_{test}$= Class label given by $N_{orig}$. 
\newline //\textit{ In case the divergence module (or the baseline node) is not present in the FALCON, the FALCON produces an error for the instance $I_{test}$. Class Label $L_{test}$= NOT FOUND and the classification process is TERMINATED at the initial node without activating other nodes.}
\State \textbf{if} $|o_{max}$ - $o_{min} | > \delta$ \textbf{then} final node ($node_f$) corresponding to the path activated by output neuron $o_{max}$ is enabled. 
\newline Class Label $L_{test}$= Class label given by final node (\textbf{$node_f$}). 
\end{algorithmic}
\end{algorithm}

In summary, the design methodology implicitly obtains the relevant features representative of the classes in the dataset and utilizes the feature consensus across classes to construct a multi-step classification tree. The divergence value $\delta$ can be adjusted during runtime to achieve the best tradeoff between accuracy and efficiency with FALCON.  We believe that the proposed approach is systematic and can be applied across all classification applications.

\section{Neuromorphic Engine: Hardware Platform for FALCON}
In this section, we describe the proposed Neuromorphic Engine (NeuE) that provides a hardware framework to execute ANNs. NeuE is a specialized many-core architecture for energy efficient processing of FALCON classification technique. NeuE delivers state-of-the art accuracy with energy efficiency by using the following two approaches: (1) hardware support for efficient data movement by spatial and temporal data reuse (FIFO, T-Buffer) to minimize the number of SRAM accesses; (2) hardware support for data gating to prevent unwanted memory reads and “Multiply and ACcumulate” (MAC) operations thereby allowing input-aware data processing. Additionally, the control unit supports selective path activation to enable FALCON.

Fig. 5 shows the block diagram of the NeuE architecture with arrows depicting the logical dataflow between the constituent units. The SRAM memory stores the input data (image pixel values and weights) for the trained neural network. Efficient data movement is achieved by buffering the input data - image data (Im) and weight data (Wt) in FIFOs and temporary output traces (T-trace) in the T-Buffer. Image data and weight data are read from SRAM memory into the FIFOs and streamed into the array of Neuron Units (NUs).  Temporary output traces computed in NUs are buffered into T-Buffer instead of being written back into the SRAM and read from the buffer when needed by the NUs for further processing. The NUs compute the product between the image data and weight data and keep accumulating it until all the inputs for a particular neuron are processed. After this, the Activation Unit (AU) processes the value in the NU and the output is returned to the SRAM.

Let's discuss the mapping of a generic neural network (fully connected) into NeuE. The neuron computations are done layer wise – read the inputs and weights from SRAM, compute all the outputs corresponding to the first layer, store back the outputs in SRAM and then proceed to the next layer. Within a layer, neurons are temporally scheduled in the NUs – the output computations for the first set of ‘N’ neurons are done. Then, the next set of ‘N’ neurons from the same layer are scheduled in the NU and the process continues until all the neurons in the current layer have been evaluated. Hence, we temporally map the different layers of the neural network and different neurons within a layer to compute the entire neural network for a given input data. Thus, NeuE is a temporally scalable architecture capable of implementing all fully connected artificial neural networks.

The logical dataflow between different components of the NeuE is also shown in Fig. 5. ‘N’ (16 in our case) input data are read from the SRAM into the Input FIFO. Each NU receives weights from its dedicated weight FIFO. Corresponding to the data in input FIFO, ‘N’ weights are read from the SRAM into each NU with each NU corresponding to a neuron. The input FIFO is flushed (new set of ‘N’ data read from and put in Input FIFO) after all the computations for the first layer neurons is done. Inputs are streamed from the input FIFO into the NU array as all the neurons in a layer share the same inputs. Once all the computations (that can be done with the current data in input FIFO) for the first set of ‘N’ neurons scheduled into the NU array is complete, the T-traces are stored into T-Buffer. The T-trace will be read back into the NU when the input FIFO gets flushed to read the new set of inputs. After, the T-trace has been written to the T-buffer, the next set of ‘N’ neurons are scheduled into the NUs, corresponding weights read from SRAM into their respective weight FIFOs and the logical flow continues as described.

Input FIFO and T-Buffer facilitate efficient data movement. Data in Input FIFO is shared by all neurons scheduled in the NUs that allows spatial reuse of input data. Additionally, temporary output traces are stored in the T-Buffer and hence allowing temporal reuse of the data in input FIFO for successive set of ‘N’ neurons in the same layer. The data in T-Buffer is also temporally reused by NUs which otherwise would be written back and fetched from the SRAM. The FALCON algorithm decomposes a bigger neural network into smaller ones thereby allowing effective T-Buffer utilization as the number of intervening trace storages before a trace buffer entry is reutilized for further accumulation are less, hence preventing them from being evicted before getting reutilized. Efficient data movement translates to $\sim$7 \% energy saving on an average across all datasets. For larger networks that cannot store all the T-traces in the T-Buffer for a layer, the T-trace is evicted and written to the SRAM memory. 

The control unit holds control registers which store information about the topology of the FALCON tree i.e. connections and size of ANNs in it. It also has the Selective-path activation unit (SAU). The SAU keeps track of network execution, gathers the outputs and selectively activates the correct path based on the output from the previous stage. Each NU is a Multiply and ACcumulate (MAC) unit. The NUs are connected in a serial fashion to allow data streaming from Input FIFO to the rightmost NU. The AU implements a piecewise linear approximation of the sigmoid function. Once, the NUs have finished the weighted summation of all inputs, the AU streams in the data from the NUs in a cyclical fashion and sends the output back to the NUs as shown in Fig. 5. Data gating is achieved by input aware weight fetching. The zero input checker disables the corresponding weight fetches for all the neurons in the layer being processed currently if the input pixel value is zero. This translates to energy saving by skipping weight reads from SRAM and corresponding multiply and accumulate computation in NUs. On an average, data gating translates to significant savings across the datasets further decreasing the overall energy consuption.


\begin{figure}[t!]
\centering
\includegraphics[width=0.5\textwidth]{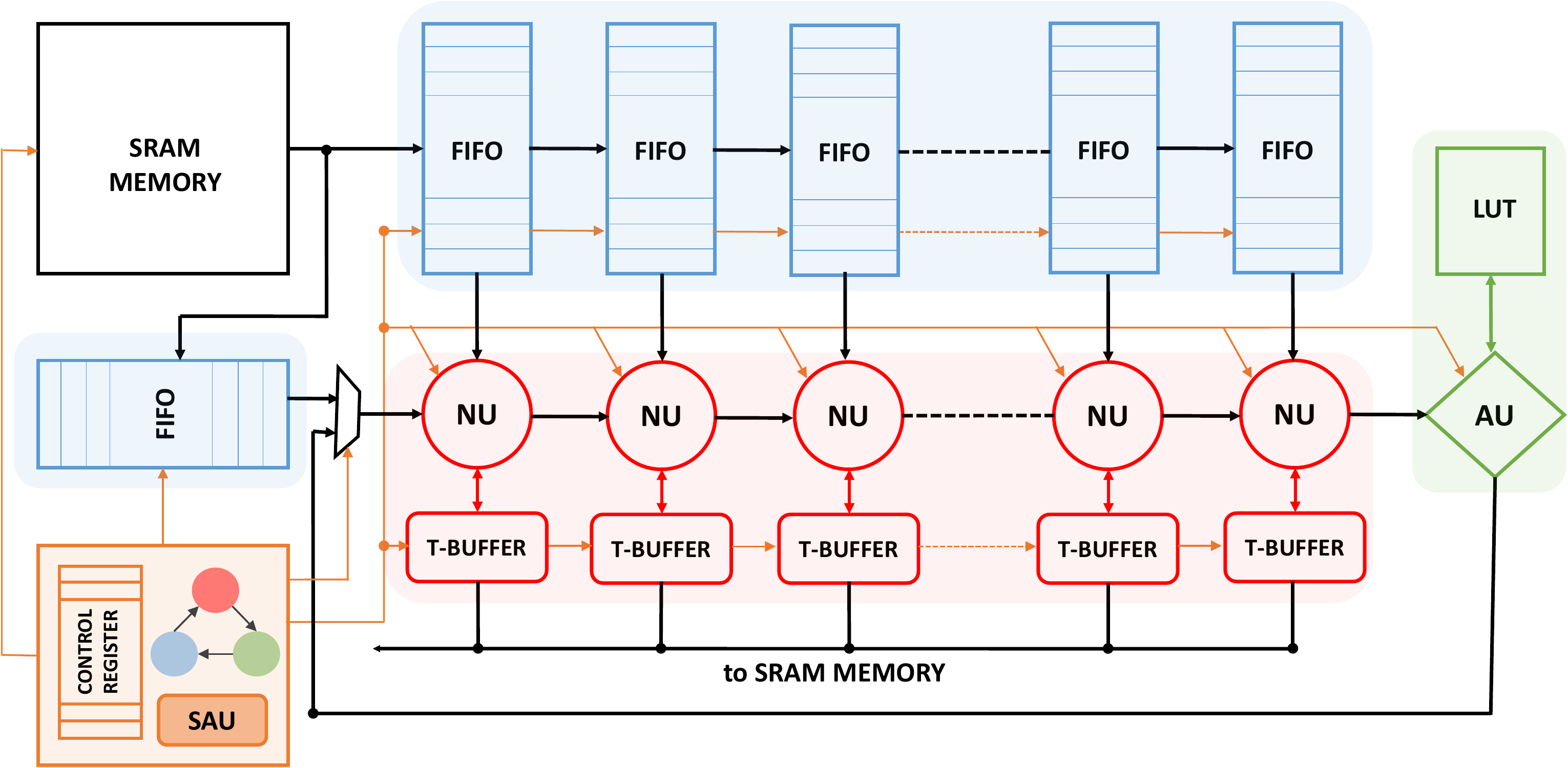}
\caption{Block Diagram of the scalable and programmable Neuromorphic Engine (NeuE). }
\end{figure}

\section{Experimental Methodology}
In this section, we describe the experimental setup used to evaluate the performance of FALCON approach. We note that our methodology is generic and can be applied to any give n-object classification task. It is apparent that images in all real-world datasets do share common features across classes which can be utilized to implement our design strategy. As an example, we have implemented a standard ANN based 12-class image recognition platform for the Caltech101 dataset \cite{fei2007learning} and 10-class platform for CIFAR10 dataset \cite{krizhevsky2010convolutional}. We have used these datasets as for our proposed methodology, the images need to be characterized with appropriate features. Caltech101/CIFAR10 have good resolution colored images that can be characterized with color/texture. For Caltech, each image is roughly around 300x200 pixels that are scaled to 75x50 pixels for hardware implementation. For CIFAR10, we used the original resolution of 32x32 pixels for evaluation. For the 12-class Caltech recognition, first we built a 4-object/8-object classifier (Fig. 6 (a, b, c)) using the design methodology discussed in Section III(C). Then, the nodes of the smaller classifiers were reused to construct a 12-object classifier as shown in Fig. 6 (d). Each node/classifier in the FALCON tree is trained using Stochastic Gradient Descent with backpropagation \cite{lecun1998gradient}. 

\begin{figure}[t!]
\centering
\includegraphics[width=0.5\textwidth]{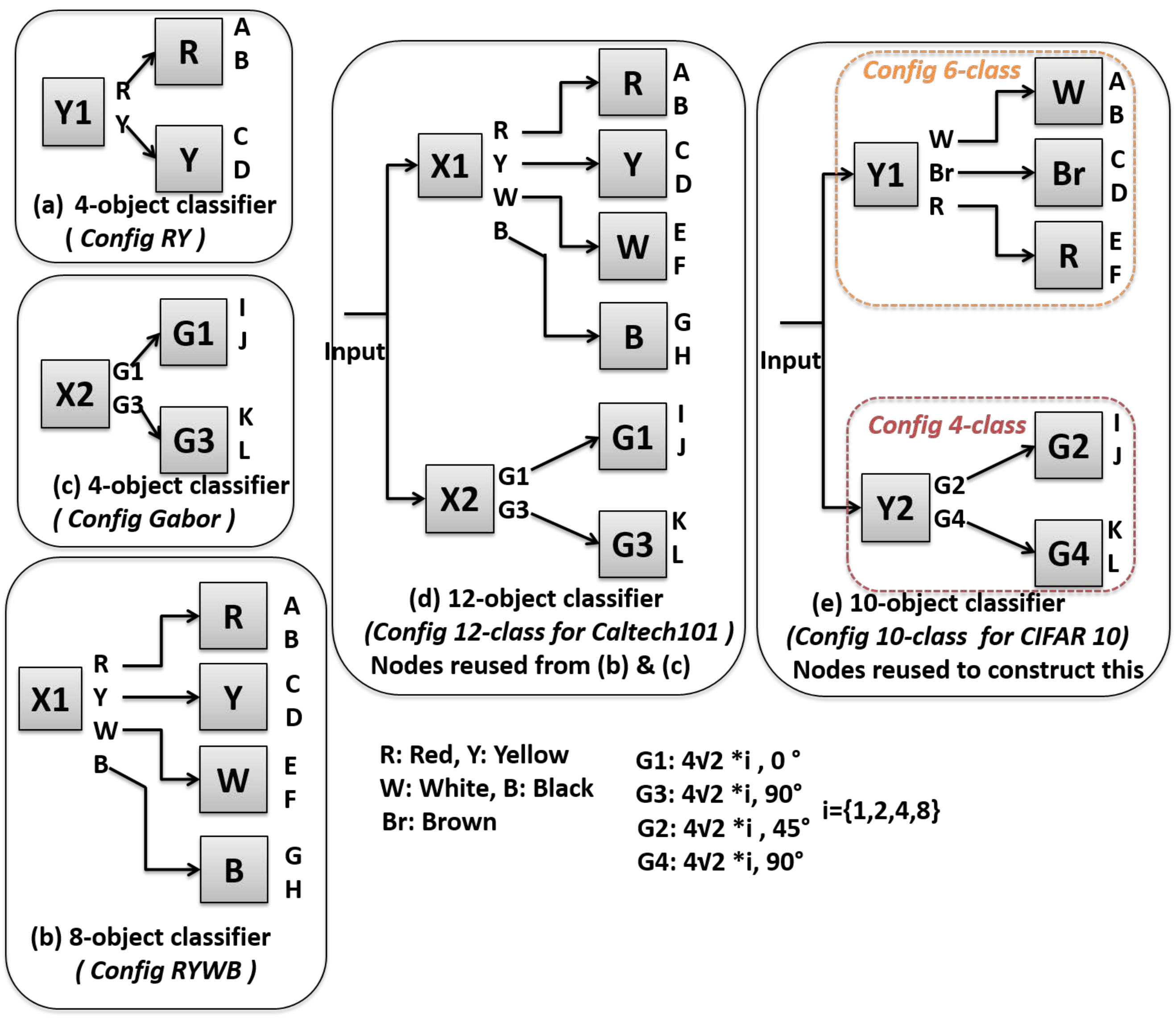}
\caption{FALCON tree configurations for n-object classification for (a-d) Caltech101 and (e) CIFAR10. }
\end{figure}

For ease of representation, the divergence module with the baseline classifier for each FALCON configuration is not shown. We can see that the initial node for each configuration is trained for different feature classes (color: Fig. 6 (a, b) and texture: Fig. 6 (c)) as deemed optimum by the feature selection methodology. R,Y,W,B are the broad color features that were obtained for classes (A - H) while G1, G3 are the texture features for classes (I-L). Please note that the nodes that were reused to build the larger classifiers (\textit{Config RYWB, Config 12-class}) did not have to be retrained at all. The FALCON shown in Fig. 6 (d) reuses the nodes in Fig. 6 (b, c) and has two initial nodes (X1, X2). During the test phase for FALCON in Fig. 6 (d) , the input image is fed to both X1, X2 and the output neuron with the maximum confidence across X1 , X2 is used to select the corresponding path to the final node. In case of the 10-class image recognition for CIFAR10, we applied the same procedure as Caltech where we built 6-object/4-object FALCON classifier configurations and reused their nodes to build the 10-object FALCON model as shown in Fig. 6 (e). For convenience in representation, we have not shown the modular representation of the smaller FALCON configurations for CIFAR10.

\begin{figure}[t!]
\centering
\includegraphics[ scale=0.5]{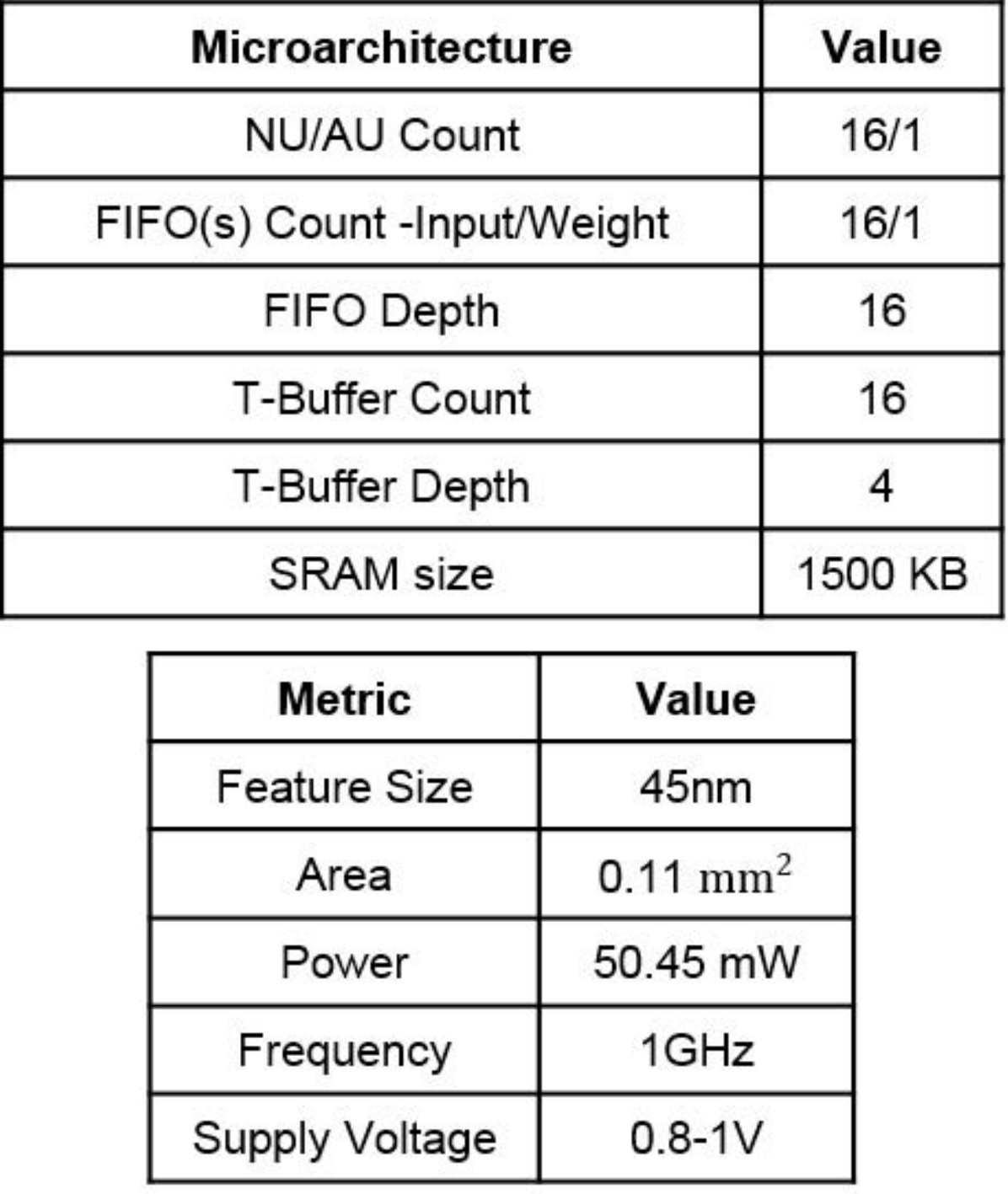}
\caption{Neuromorphic Engine (NeuE) parameters and metrics}
\end{figure}

For hardware implementation, we implemented the NeuE at the Register-Transfer-Level (RTL) and mapped to the IBM 45nm technology using Synopsys Design Compiler. We used Synopsys Power Compiler to estimate energy consumption of the implementation. The key micro-architectural parameters and implementation metrics for the core of the NeuE are shown in Fig. 7. Each of the configurations in Fig. 6 for Caltech101 and CIFAR10 were ported manually to the NeuE platform and the baseline (corresponding single NN classifier for each FALCON Config in Fig. 6) was well optimized for energy. The NeuE operates at 1GHz core clock resulting in an average total power consumption of 72.68 mW across the 12-class Caltech/10-class CIFAR recognition implementations. The execution core and the memory consume 78.92\% and 21.07\% of the total power, respectively. To minimize leakage power and better optimize the energy of baseline classifiers for fare comparison with FALCON, we used a supply voltage of 0.8V for memory and that of 1V for execution core operation in the NeuE. For runtime analysis, we implemented each of the configurations of Fig. 6 in Matlab and measured runtime for the applications using performance counters on Intel Core i7 3.60 GHz processor with 16 GB RAM. Please note that the software baseline implementation was aggressively optimized for performance.

\section{Results}
In this section, we present the experimental results that demonstrate the benefits of our approach. We use Caltech101 as our primary benchmark to evaluate the benefits with selective classification.

\subsection{Energy Improvement}
\begin{figure}[t!]
\centering
\includegraphics[ scale= 0.55]{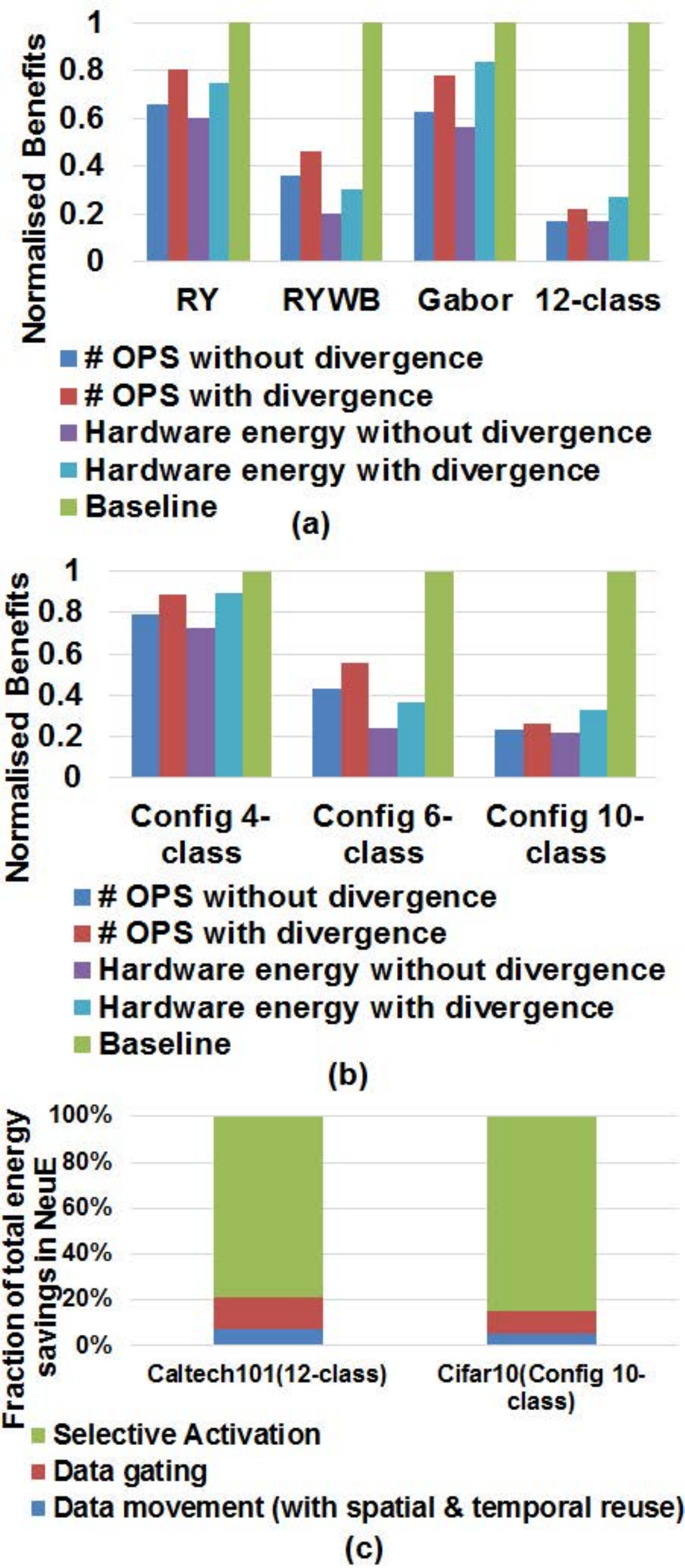}
\caption{Normalized benefits (OPS/energy) for each FALCON configuration (in Fig. 6) with or without divergence with respect to baseline (a) for Caltech101 dataset (b) for CIFAR10 dataset. (c) Fraction of energy savings due to various architectural techniques observed for different datasets in NeuE }
\end{figure}

Fig. 8 (a) shows the improvement in efficiency with respect to the traditional single NN classifier (which forms the baseline) for each configuration of Fig. 6 (a-d) with and without the divergence module for Caltech101. We quantify efficiency in terms of two metrics: (i) the average number of operations (or MAC computations) per input (OPS), (ii) energy of hardware implementation on NeuE. The OPS and energy of each FALCON \textit{Config} is normalized to a NeuE implementation of the corresponding baseline classifier. Note that this is already a highly optimized baseline since the NeuE architecture is customized to the characteristics of ANNs. We observe that while our proposed FALCON approach yields 1.51x-5.97x (average: 3.74x) improvement in average OPS/input compared to the baseline in the case without divergence, the benefits are slightly lower 1.24x-4.59x (average: 2.92x) with divergence. This is obvious because the baseline classifier is not present as a final node in the FALCON tree in the case without divergence. It is clearly seen in Fig. 8 (a) that the benefits observed increases by almost 1.5x each time we scale up from a 4-object classification (\textit{Config RY, Gabor}) to an 8- object (\textit{Config RYWB}) /12-object (\textit{Config 12-class}) problem. This can be attributed to the fact that the complexity of the baseline classifier increases substantially in order to get a reasonable classification accuracy for a given n-object classification problem. In contrast, FALCON invokes multi-step classification based on feature information in the input data. Thus, the decomposition of the classification problem into simpler tasks allows us to use a cluster of less complex nodes (with lower dimensional feature vector as input to final nodes) that combined with selective activation yields larger benefits. Additionally, the reuse of nodes contributes further to the increased benefits while scaling up from small to larger classification problems. Please note that the benefits shown include the additional cost of HSV and Gabor filtering for the FALCON implementation. In case of hardware execution on NeuE, the energy improvements obtained are 3.66x/5.91x for the 12-object classification with/without divergence respectively as illustrated in Fig. 8 (a). Similarly, Fig. 8 (b) shows the normalized benefits (OPS and energy) observed for the FALCON implementation of CIFAR10 with the three configurations from Fig. 6 (e). On an average, FALCON achieves 3.05x/4.55x improvement in energy and 3.82x/4.26x improvement in OPS with \textit{Config 10-class} (Fig. 6 (e)) for 10-object classification. 

We also show the fraction of total energy savings observed in the hardware platform NeuE due to other standard architectural design techniques besides selective activation for each of the datasets (Caltech101, CIFAR10) in Fig. 8(c). It is clearly seen that while data gating and data movement techniques provide $\sim$20\% of the total savings in each case, the majority of savings is observed due to FALCON methodology that invokes selective activation. A noteworthy obsevation here is that data gating/movement provides more benefits for Caltech101 than CIFAR10. This can be attributed to the fact that input size dimensions for Caltech101 (75x50) is greater than CIFAR10 (32x32) that results in more near-zero pixels for the former and thus more data gating. Also, in Caltech101 (Fig. 6 (d)) the number of decomposed classifiers obtained from FALCON is greater than that of CIFAR10 (Fig. 6(e)). The T-Buffer reutilization is more in the former case resulting in larger \% of savings due to efficient data movement than the latter.

Fig. 9 shows the normalized accuracy of each configuration in Fig. 6 (a-d) for Caltech101 with/without the divergence module with respect to the corresponding baseline classifier. The accuracies of the FALCON \textit{Configs} are normalized with respect to the corresponding baseline. For example, the accuracy of the baseline for the 12-class problem is 94.2\% that is set to 1 and the corresponding FALCON (\textit{Config 12-class}) is normalized against it.  It is evident that while the configuration with divergence module yields iso-accuracy as that of the baseline, the absence of the module results in a decline in accuracy by 1.7\%-3.9\%. For CIFAR10, the FALCON \textit{Config 10-class} yields a 2.8\% accuracy decline without the divergence module with respect to the baseline (with absolute accuracy of 78.8\%) for the 10-class recognition problem. As discussed in Section III (B.2), this degradation is due to the errors given out at the initial node for those test instances that have more than one feature as representative information. However, for hardware implementations where energy-efficiency is crucial, 2-4\% decline in accuracy may be permissible. Note that the test speed efficiency that is dependent upon the number of evaluated classifiers and the complexity of each classifier is similar to the savings as observed from OPS/input calculation.

\begin{figure}[t!]
\centering
\includegraphics[ scale= 0.45]{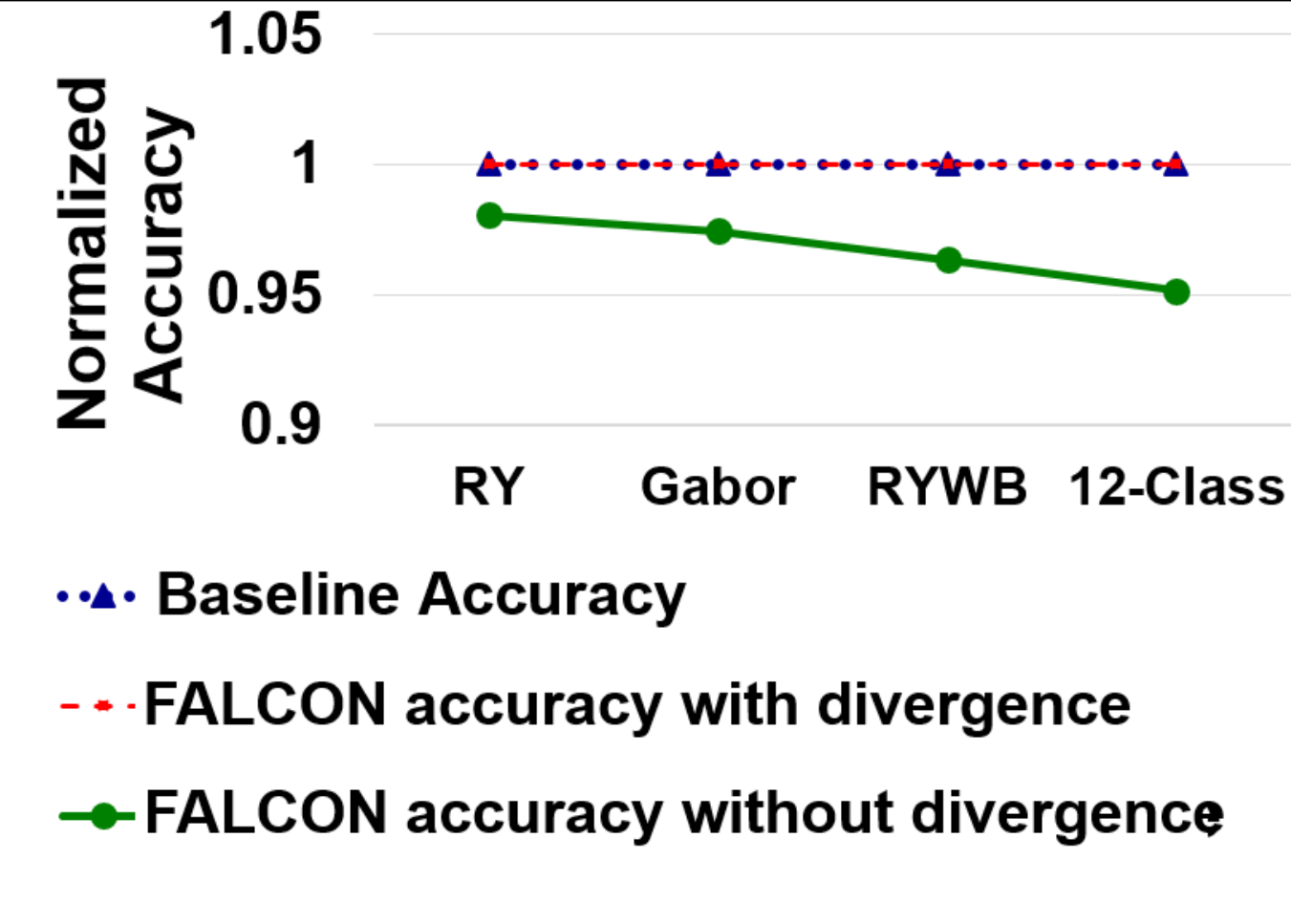}
\caption{Normalized Accuracy for each FALCON configuration for Caltech101 dataset}
\end{figure}

\subsection{Improvement in training time}
One of the big challenges in machine learning is the time needed to train neural networks to obtain a reasonable accuracy for large classification tasks. In fact, the software based implementation of large scale problems require accelerators like GPUs that use CUDA for faster and high performance neural network training \cite{ngiam2011optimization}. Since feature based classification enables the nodes in the FALCON tree to be trained for simpler tasks, we can conjecture that there should be reduction in training time with FALCON. For example, referring to Fig. 6 (b), $Config RYWB$ is originally an 8-object classifier decomposed into a 4-object (initial node $X1$) and cluster of small 2-object classifiers (node $R, Y, W, B$). Hence, these nodes will converge to the global error minima much faster than the baseline classifier. However, it is understood from the design methodology that prior to constructing the FALCON tree, the feature selection methodology has to be invoked to obtain the appropriate feature vectors. This would add extra overhead on the training time. Fig. 10 illustrates the normalized training time observed for each configuration of Fig. 6 (without the divergence module) with respect to the baseline. 

\begin{figure}[t!]
\centering
\includegraphics[ scale= 0.5]{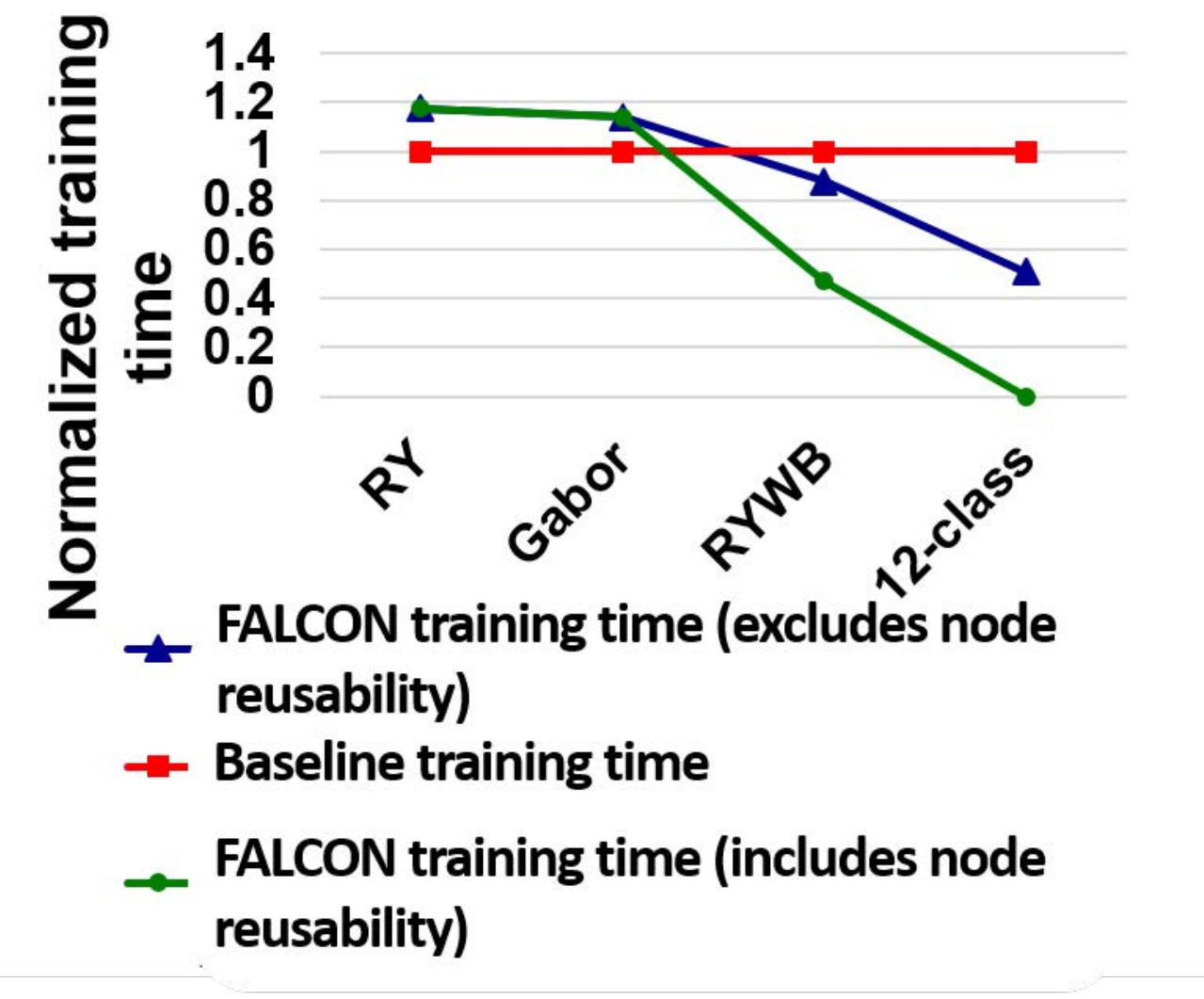}
\caption{Improvement in training time in SSC with/without node reusability}
\end{figure}

The additional overhead of feature selection is more pronounced for smaller tasks ($Config RY/Gabor$) due to which the time for training the FALCON in these cases is slightly more (1.17x/1.13x) than that of the baseline. However, as we scale to larger problems, we observe that there is a significant improvement (1.14x for Config RYWB/ 1.96x for Config 12-class) in training time with FALCON even when node reusability is not taken into account. This is because the baseline classifier becomes increasingly complex and difficult to train for complex tasks. In contrast, FALCON in spite of the overhead trains easily because of problem decomposition. Now, if we take into account node reusability, then, scaling up the problem from $Config RY$ (4-object) to $Config RYWB$ (8-object) doesn’t require training of the nodes $R$ and $Y$. Thus, reuse of nodes will cause the training time to further reduce that is evident in Fig. 10. Since the 12-object FALCON ($Config 12-class$) is built reusing the nodes from $Config RYWB$ and $Config Gabor$, it should ideally require no extra training time that is seen from Fig. 10. It is very evident that with FALCON, the classifier architecture is optimized such that it can be easily mapped to GPU/ CUDA framework, in software simulations, giving ultra-high performance on enormous datasets. This shows the effectiveness of FALCON.

\subsection{Efficiency-Accuracy tradeoff using divergence $\delta$}
The divergence module discussed in Section III (B.2) enables the baseline node in the FALCON tree depending upon the divergence value, $\delta$, set by the user. Fig. 11 shows the variation in normalized energy (with respect to baseline) and the accuracy for the FALCON (\textit{Config RY} in Fig. 6(a)) with different $\delta$. Setting $\delta$ to a low value implies that the baseline node will be activated few times and more inputs will be passed to the final nodes (Node \textit{R, Y}: Fig. 6 (a)) for classification. Thus, initially we observe more reduction in energy as compared to the baseline. 

\begin{figure}[t!]
\centering
\includegraphics[ scale= 0.55]{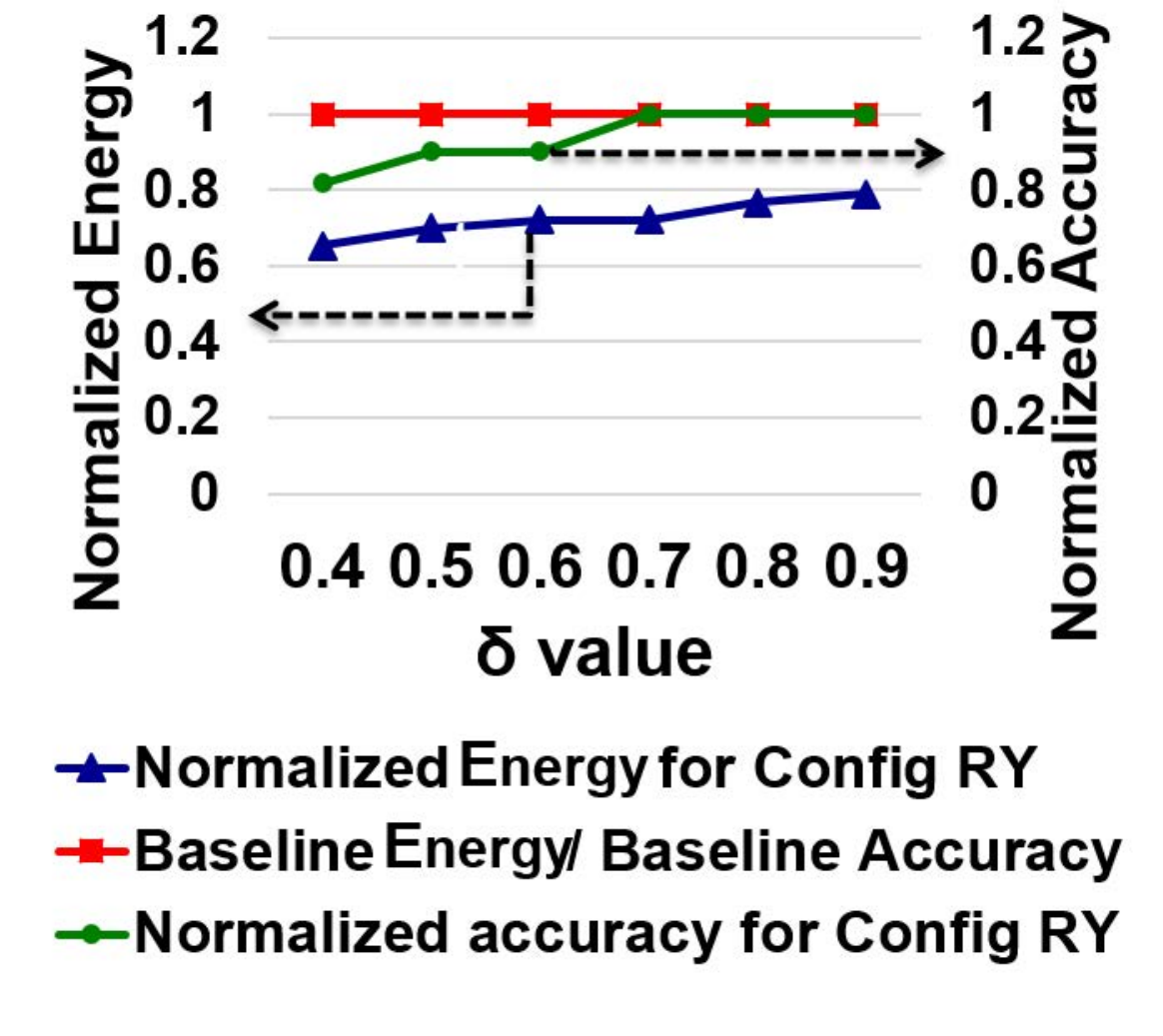}
\caption{Efficiency vs. accuracy using divergence $\delta$}
\end{figure}

However, in such cases, the difference between the confidences at the output neurons of the initial node (Node \textit{Y1}) is also low. There is a high probability that the initial node does not activate the final nodes accurately i.e. it wrongly activates the path to final node \textit{R} when the test instance originally should be classified by node \textit{Y}. Thus, we see that the accuracy of the FALCON is lower than that of the baseline. Increasing $\delta$ improves accuracy at the cost of increase in energy as the baseline is enabled more now. However, beyond a particular $\delta$, the FALCON achieves iso-accuracy with that baseline. This value of $\delta$ corresponds to the maximum efficiency that can be achieved for the given FALCON configuration. In Fig. 11, we observe that iso-accuracy is attained for $\delta$ = 0.7. The energy would still continue to increase beyond this point. So, we can regulate $\delta$ during runtime to trade accuracy for efficiency. 

\subsection{Adding new nodes to FALCON tree}
Till now, we have discussed reusing nodes from smaller classification tasks to scale up to larger problems when the new classes have different feature information (like \textit{Config RY} to \textit{Config RYWB} in Fig. 6 required incorporating classes with features white, black). Consider a case where we need to extend the Config RY in Fig. 6 (a) to incorporate new classes that have red as a representative feature. In this case, we need to retrain Node \textit{Y1} (Fig. 6 (a)) with the additional classes and also modify the final node corresponding to the path activated by R. Hence, we have two options as shown in Fig. 12 (b): i) Retrain the final node R with new classes (Config Retrain) and ii) Add a new node (Node \textit{R’}) to the path (\textit{Config New}). However, the option that gives the maximum benefits depends on the number of new classes to be added. 

\begin{figure}[t!]
\centering
\includegraphics[scale=0.35]{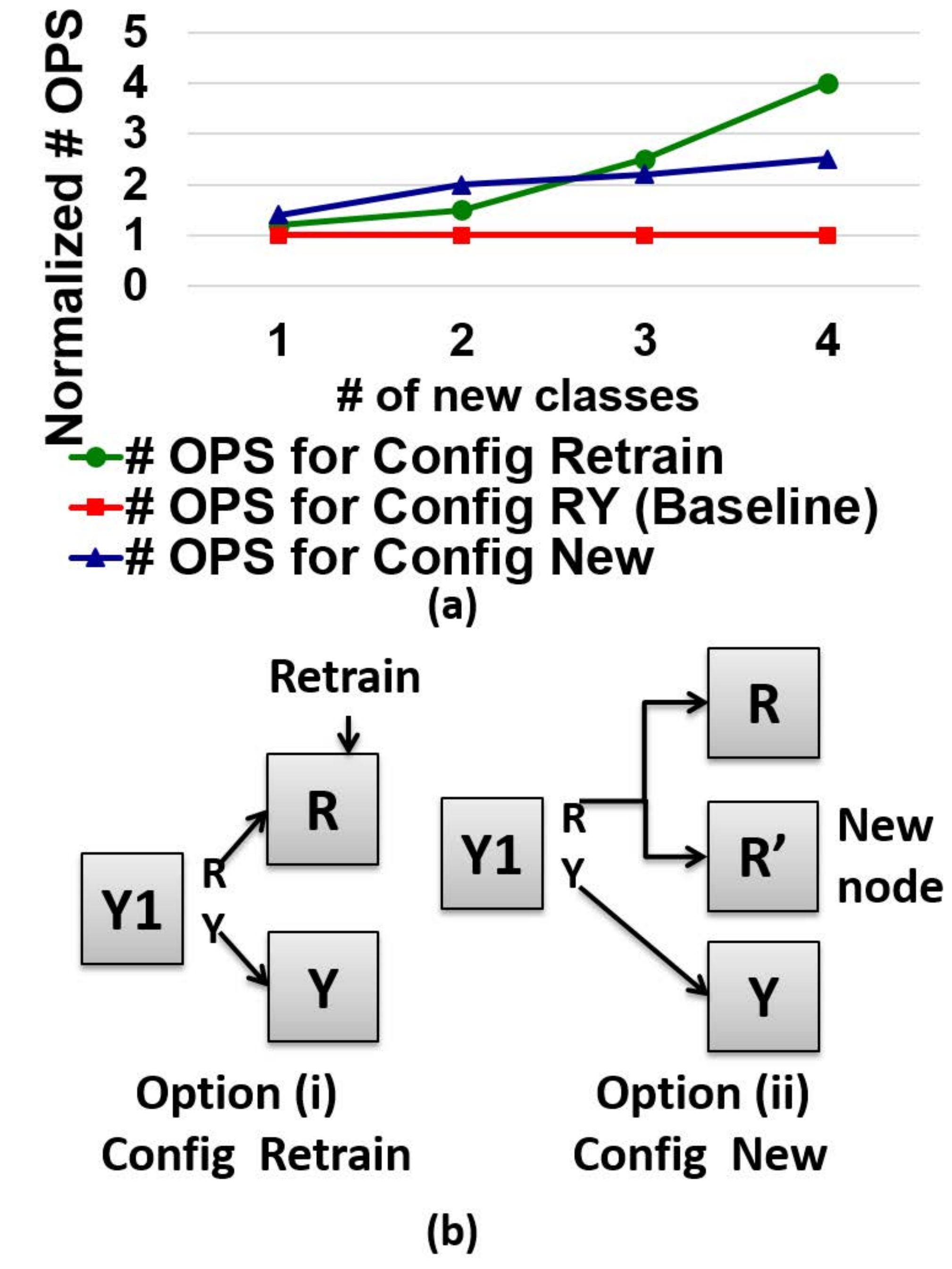}
\caption{(a) Normalized \# OPS for \textit{Config Retrain} and \textit{New} with increase in number of new classes (b) Structure of \textit{Config Retrain} and \textit{Config New}}
\end{figure}

Fig. 12 (a) shows the normalized OPS (that also quantifies efficiency) for both options as the number of new classes (to be added) is increased. It is evident that both \textit{Config New} and \textit{Retrain} will have higher \#OPS than the initial \textit{Config RY} (which forms the Baseline here) due to the presence of new classes. In option (i), addition of a new node implies that both nodes (\textit{R, R’}) have to be activated to obtain the final classification result. In contrast, with option (ii), only the retrained node \textit{R} needs to be enabled. Thus, as long as the complexity of retrained node \textit{R} in \textit{Config Retrain} is less than the combined complexity of Node \textit{R} and \textit{R’} in \textit{Config New}, option (i) yields more computational benefits. Thus, initially we observe higher \# OPS with \textit{Config New}. However, as we increase the number of new classes, the complexity of retrained Node \textit{R} also increases in order to maintain competitive classification accuracy. At some point, this complexity would overcome the cost penalty that activating two nodes (\textit{R, R’}) imposes. Beyond this point, option (ii) yields more benefits. In Fig. 12 (b), for \# of new classes $>$ 2, adding new nodes is preferred. This behavior is taken into account while constructing the FALCON tree to get maximum savings. A similar analysis was done to construct \textit{Config RYWB} (Fig. 6 (b)) with a single initial node (\textit{X1}) as opposed to multiple initial nodes. \textit{Config 12-class} (Fig. 6 (d)) also has two initial nodes \textit{X1, X2} due to the given analysis.

\section{Comparison of FALCON with Deep Learning Networks}
Deep Learning Networks (DLNs) are the current state-of-the-art classifier models that have demonstrated remarkable performance on computer vision and related applications. While these large-scale networks are very powerful, they consume considerable storage and computational resources. The proposed FALCON methodology uses the characteristic features of images to train simple classifier models with lower complexity for efficient classification. As a way of determining the effiectiveness of our proposed methodology with state-of-the-art methods, we compare FALCON with deep learning models and gauge the energy vs. accuracy tradeoff obtained from both the models. We chose two deep learning models of different depths (or layers), namely, ConvNet with 5 layers \cite{convnetJS} and Wide ResNet \cite{zagoruyko2016wide} with 40 layers (and a widening factor of 2) for efficiency comparison with FALCON methodology on the CIFAR-10 dataset. Please note that feedforward ANNs are used as the baseline as well as the classifier nodes of the FALCON tree. As a result, the accuracy that can be obtained with such networks is generally low as compared to that of several layered DLNs. Hence, for fair comparison of accuracy and energy benefits, we compare our proposed FALCON configuration with the above deep learning networks, ConvNet that yields iso-accuracy ($\sim$78.8\%) as that of FALCON and Wide ResNet that yields an improved accuracy of $\sim$93.3\%. It is evident that the ConvNet architecture owing to the shallow depth achieves lower accuracy than that of Wide ResNet. 

\begin{figure}[t!]
\centering
\includegraphics[scale=0.4]{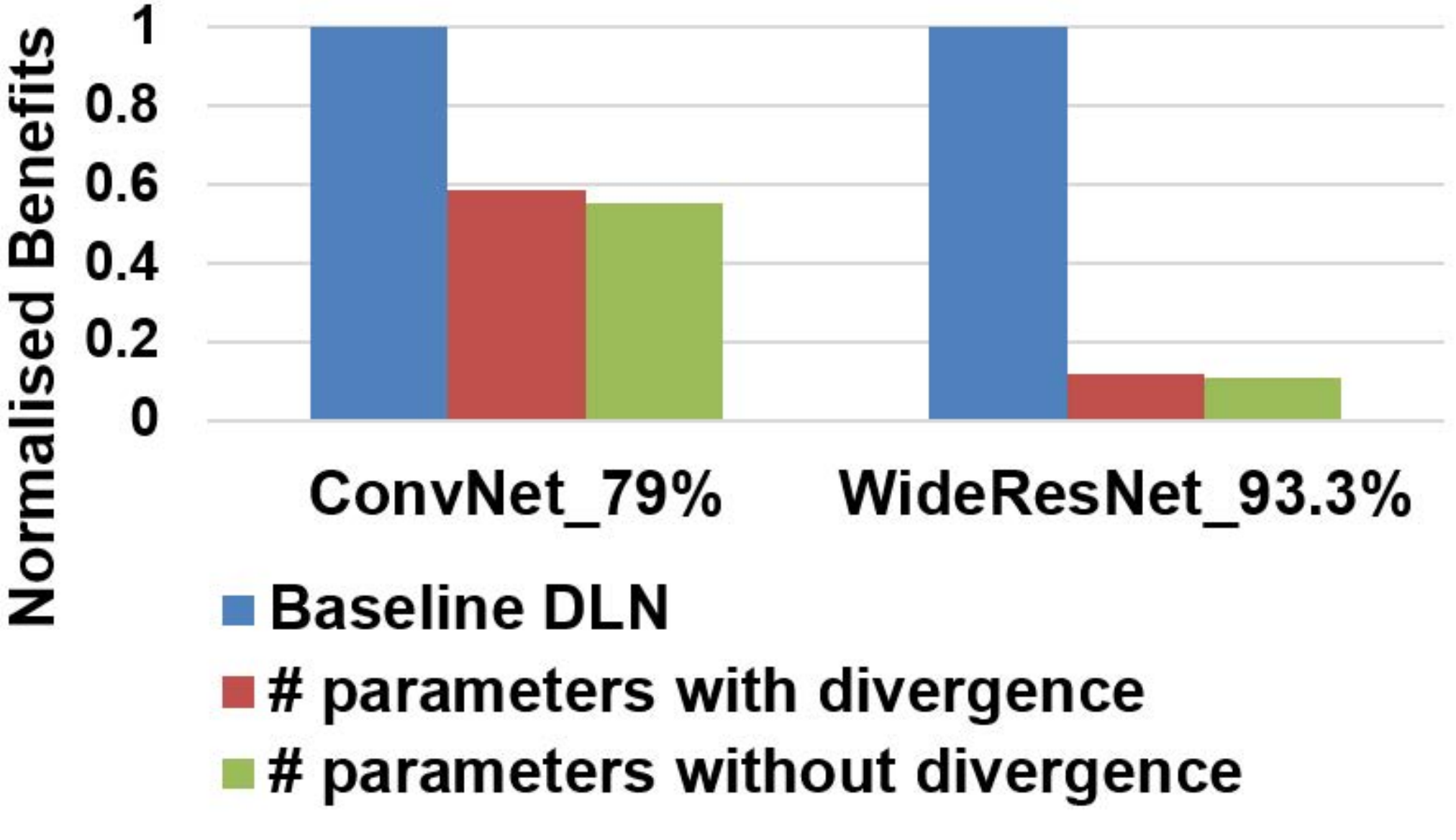}
\caption{Normalized benefits for \textit{Config 10-class} FALCON classifier with or without divergence module with respect to Deep Learning Networks with varying depths and accuracy: ConvNet (78.8\% accuracy), Wide ResNet (93.3\%accuracy) }
\end{figure}

Fig. 13 shows the normalized benefits observed with FALCON \textit{Config 10-class} for CIFAR-10 (refer Fig. 6 (e)), with and without the divergence module, as compared to the ConvNet and WideResNet DLN models that forms the baseline. It is worth mentioning that we use the total number of parameters or weights to quantify the computational complexity (or efficiency) in this case. In fact, many recent works \cite{zagoruyko2016wide, han2015learning} have used this metric to quantify the computational benefits. Thus, the total number of parameters (that directly translates to energy consumption of a model) serves as an objective metric for efficiency comparison of FALCON with DLNs. It is clearly seen from Fig. 13 that FALCON yields improved benefits as compared to both the DLNs. While the benefits observed are $\sim$1.71x/1.82x (with/without divergence) with respect to ConvNet, the improvement increases significantly to $\sim$8.7x/9.3x as compared to Wide ResNet model. Thus, we can infer that FALCON owing to selective activation yields significant computational savings as compared to DLNs and is very efficient to deploy on hardware. For the shallow ConvNet that yields lower accuracy, FALCON tends to be the energy-efficient choice while achieving similar output quality.  However, the accuracy obtained with FALCON is $\sim$14\% lower than that of Wide ResNet. Hence, DLNs that consist of multiple layers, though being highly computationally intensive than FALCON, will outperform in terms of accuracy. 

Please note, the shallow ConvNet model was implemented in the widely used Torch platform \cite{collobert2002torch} to train and test CIFAR-10 (with similar architecture and number of parameters as in \cite{convnetJS}). The accuracy and parameters for the Wide ResNet model are directly taken from \cite{zagoruyko2016wide}. 

\section{Conclusion}
In this paper, we propose FALCON: Feature Driven Selective Classification, based on the biological visual recognition process, for energy-efficient realization of neural networks for multi-object classification. We utilize the feature similarity (or concensus) across multiple classes of images in a real-world dataset to break down the classification problem into simpler tasks organized in a tree-fashion. We developed a systematic methodology to select the appropriate features (color and texture for images) and construct the FALCON tree for a given n-object classification task. The structure of FALCON provides us with a significant advantage of reusing tree nodes from smaller classification tasks to implement large-scale problems thereby contributing to the reduction in training time as we scale to larger tasks. FALCON invokes selective activation of only those nodes and branches relevant to a particular input, while keeping remaining nodes idle, resulting in an energy-efficient classification process. 

It is noteworthy to mention that the current FALCON methdology employs a feature selction process that clusters classes at the root node based on a single feature similar to a group of classes. Thus, we add the divergence module (or baseline classifier as an extra node) to maintain the accuracy of the FALCON tree for those classes that have more than one feature in common. For such cases (with divergence module), we observe lesser energy benefits. A feature selection algorithm that  searches for more distinctive features similar across classes will prevent the use of the divergence module, yielding higher energy savings while maintaining iso-accuracy with that of baseline. Furthermore, while the computational benefits from the proposed tree-based FALCON technique is evident, in order to match the high output quality observed with state-of-the-art deep learning models, we might have to employ better feature selection techniques that will be able to give optimal features for the initial nodes of the hierarchy. Recent works have proposed partitioning techniques that provide competetive classification even for large-scale problems \cite{yuan2006automatic,bengio2010label,gao2011discriminative}. Hence, further research can be done to improvise the feature selection process.

\section*{Acknowledgment}
This work was supported in part by C-SPIN, one of the six centers of StarNet, a Semiconductor Research Corporation Program, sponsored by MARCO and DARPA, by the Semiconductor Research Corporation, the National Science Foundation, Intel Corporation and by the Vannevar Bush Faculty Fellowship.

\ifCLASSOPTIONcaptionsoff
  \newpage
\fi


\begin{thebibliography}{10}
\providecommand{\url}[1]{#1}
\csname url@samestyle\endcsname
\providecommand{\newblock}{\relax}
\providecommand{\bibinfo}[2]{#2}
\providecommand{\BIBentrySTDinterwordspacing}{\spaceskip=0pt\relax}
\providecommand{\BIBentryALTinterwordstretchfactor}{4}
\providecommand{\BIBentryALTinterwordspacing}{\spaceskip=\fontdimen2\font plus
\BIBentryALTinterwordstretchfactor\fontdimen3\font minus
  \fontdimen4\font\relax}
\providecommand{\BIBforeignlanguage}[2]{{%
\expandafter\ifx\csname l@#1\endcsname\relax
\typeout{** WARNING: IEEEtran.bst: No hyphenation pattern has been}%
\typeout{** loaded for the language `#1'. Using the pattern for}%
\typeout{** the default language instead.}%
\else
\language=\csname l@#1\endcsname
\fi
#2}}
\providecommand{\BIBdecl}{\relax}
\BIBdecl

\bibitem{dubey2005recognition}
P.~Dubey, ``Recognition, mining and synthesis moves computers to the era of
  tera,'' \emph{Technology@ Intel Magazine}, vol.~9, no.~2, pp. 1--10, 2005.

\bibitem{jones2014learning}
N.~Jones \emph{et~al.}, ``The learning machines,'' \emph{Nature}, vol. 505, no.
  7482, pp. 146--148, 2014.

\bibitem{netzer2011reading}
Y.~Netzer, T.~Wang, A.~Coates, A.~Bissacco, B.~Wu, and A.~Y. Ng, ``Reading
  digits in natural images with unsupervised feature learning,'' 2011.

\bibitem{krizhevsky2012imagenet}
A.~Krizhevsky, I.~Sutskever, and G.~E. Hinton, ``Imagenet classification with
  deep convolutional neural networks,'' in \emph{Advances in neural information
  processing systems}, 2012, pp. 1097--1105.

\bibitem{ramasubramanian2014spindle}
S.~G. Ramasubramanian, R.~Venkatesan, M.~Sharad, K.~Roy, and A.~Raghunathan,
  ``Spindle: Spintronic deep learning engine for large-scale neuromorphic
  computing,'' in \emph{Proceedings of the 2014 international symposium on Low
  power electronics and design}.\hskip 1em plus 0.5em minus 0.4em\relax ACM,
  2014, pp. 15--20.

\bibitem{whitney2009neuroscience}
D.~Whitney, ``Neuroscience: toward unbinding the binding problem,''
  \emph{Current Biology}, vol.~19, no.~6, pp. R251--R253, 2009.

\bibitem{ungerleider2000mechanisms}
S.~K. Ungerleider and L.~G, ``Mechanisms of visual attention in the human
  cortex,'' \emph{Annual review of neuroscience}, vol.~23, no.~1, pp. 315--341,
  2000.

\bibitem{deng2014ensemble}
L.~Deng and J.~C. Platt, ``Ensemble deep learning for speech recognition.'' in
  \emph{INTERSPEECH}, 2014, pp. 1915--1919.

\bibitem{sun2013deep}
Y.~Sun, X.~Wang, and X.~Tang, ``Deep convolutional network cascade for facial
  point detection,'' in \emph{Proceedings of the IEEE Conference on Computer
  Vision and Pattern Recognition}, 2013, pp. 3476--3483.

\bibitem{liu2007survey}
Y.~Liu, D.~Zhang, G.~Lu, and W.-Y. Ma, ``A survey of content-based image
  retrieval with high-level semantics,'' \emph{Pattern recognition}, vol.~40,
  no.~1, pp. 262--282, 2007.

\bibitem{smeulders2000content}
A.~W. Smeulders, M.~Worring, S.~Santini, A.~Gupta, and R.~Jain, ``Content-based
  image retrieval at the end of the early years,'' \emph{IEEE Transactions on
  pattern analysis and machine intelligence}, vol.~22, no.~12, pp. 1349--1380,
  2000.

\bibitem{allwein2000reducing}
E.~L. Allwein, R.~E. Schapire, and Y.~Singer, ``Reducing multiclass to binary:
  A unifying approach for margin classifiers,'' \emph{Journal of machine
  learning research}, vol.~1, no. Dec, pp. 113--141, 2000.

\bibitem{torralba2004sharing}
A.~Torralba, K.~P. Murphy, and W.~T. Freeman, ``Sharing features: efficient
  boosting procedures for multiclass object detection,'' in \emph{Computer
  Vision and Pattern Recognition, 2004. CVPR 2004. Proceedings of the 2004 IEEE
  Computer Society Conference on}, vol.~2.\hskip 1em plus 0.5em minus
  0.4em\relax IEEE, 2004, pp. II--762.

\bibitem{bengio2010label}
S.~Bengio, J.~Weston, and D.~Grangier, ``Label embedding trees for large
  multi-class tasks,'' in \emph{Advances in Neural Information Processing
  Systems}, 2010, pp. 163--171.

\bibitem{deng2011fast}
J.~Deng, S.~Satheesh, A.~C. Berg, and F.~Li, ``Fast and balanced: Efficient
  label tree learning for large scale object recognition,'' in \emph{Advances
  in Neural Information Processing Systems}, 2011, pp. 567--575.

\bibitem{beygelzimer2009conditional}
A.~Beygelzimer, J.~Langford, Y.~Lifshits, G.~Sorkin, and A.~Strehl,
  ``Conditional probability tree estimation analysis and algorithms,'' in
  \emph{Proceedings of the Twenty-Fifth Conference on Uncertainty in Artificial
  Intelligence}.\hskip 1em plus 0.5em minus 0.4em\relax AUAI Press, 2009, pp.
  51--58.

\bibitem{rastegari2012attribute}
M.~Rastegari, A.~Farhadi, and D.~Forsyth, ``Attribute discovery via predictable
  discriminative binary codes,'' in \emph{European Conference on Computer
  Vision}.\hskip 1em plus 0.5em minus 0.4em\relax Springer, 2012, pp. 876--889.

\bibitem{zweig2007exploiting}
A.~Zweig and D.~Weinshall, ``Exploiting object hierarchy: Combining models from
  different category levels,'' in \emph{2007 IEEE 11th International Conference
  on Computer Vision}.\hskip 1em plus 0.5em minus 0.4em\relax IEEE, 2007, pp.
  1--8.

\bibitem{panda2016invited}
P.~Panda, A.~Sengupta, S.~S. Sarwar, G.~Srinivasan, S.~Venkataramani,
  A.~Raghunathan, and K.~Roy, ``Invited-cross-layer approximations for
  neuromorphic computing: from devices to circuits and systems,'' in
  \emph{Proceedings of the 53rd Annual Design Automation Conference}.\hskip 1em
  plus 0.5em minus 0.4em\relax ACM, 2016, p.~98.

\bibitem{venkataramani2014axnn}
S.~Venkataramani, A.~Ranjan, K.~Roy, and A.~Raghunathan, ``Axnn:
  energy-efficient neuromorphic systems using approximate computing,'' in
  \emph{Proceedings of the 2014 international symposium on Low power
  electronics and design}.\hskip 1em plus 0.5em minus 0.4em\relax ACM, 2014,
  pp. 27--32.

\bibitem{panda2015object}
P.~Panda, A.~Sengupta, S.~Venkataramani, A.~Raghunathan, and K.~Roy, ``Object
  detection using semantic decomposition for energy-efficient neural
  computing,'' \emph{arXiv preprint arXiv:1509.08970}, 2015.

\bibitem{panda2017energy}
P.~Panda, A.~Sengupta, and K.~Roy, ``Energy-efficient and improved image
  recognition with conditional deep learning,'' \emph{ACM Journal on Emerging
  Technologies in Computing Systems (JETC)}, vol.~13, no.~3, p.~33, 2017.

\bibitem{chakradhar2010dynamically}
S.~Chakradhar, M.~Sankaradas, V.~Jakkula, and S.~Cadambi, ``A dynamically
  configurable coprocessor for convolutional neural networks,'' in \emph{ACM
  SIGARCH Computer Architecture News}, vol.~38, no.~3.\hskip 1em plus 0.5em
  minus 0.4em\relax ACM, 2010, pp. 247--257.

\bibitem{chen201614}
Y.-H. Chen, T.~Krishna, J.~Emer, and V.~Sze, ``14.5 eyeriss: An
  energy-efficient reconfigurable accelerator for deep convolutional neural
  networks,'' in \emph{2016 IEEE International Solid-State Circuits Conference
  (ISSCC)}.\hskip 1em plus 0.5em minus 0.4em\relax IEEE, 2016, pp. 262--263.

\bibitem{ngiam2011optimization}
J.~Ngiam, A.~Coates, A.~Lahiri, B.~Prochnow, Q.~V. Le, and A.~Y. Ng, ``On
  optimization methods for deep learning,'' in \emph{Proceedings of the 28th
  International Conference on Machine Learning (ICML-11)}, 2011, pp. 265--272.

\bibitem{rajendran2013specifications}
B.~Rajendran, Y.~Liu, J.-s. Seo, K.~Gopalakrishnan, L.~Chang, D.~J. Friedman,
  and M.~B. Ritter, ``Specifications of nanoscale devices and circuits for
  neuromorphic computational systems,'' \emph{IEEE Transactions on Electron
  Devices}, vol.~60, no.~1, pp. 246--253, 2013.

\bibitem{jo2010nanoscale}
S.~H. Jo, T.~Chang, I.~Ebong, B.~B. Bhadviya, P.~Mazumder, and W.~Lu,
  ``Nanoscale memristor device as synapse in neuromorphic systems,'' \emph{Nano
  letters}, vol.~10, no.~4, pp. 1297--1301, 2010.

\bibitem{roy2013beyond}
K.~Roy, M.~Sharad, D.~Fan, and K.~Yogendra, ``Beyond charge-based computation:
  Boolean and non-boolean computing with spin torque devices,'' in \emph{Low
  Power Electronics and Design (ISLPED), 2013 IEEE International Symposium
  on}.\hskip 1em plus 0.5em minus 0.4em\relax IEEE, 2013, pp. 139--142.

\bibitem{singha2012content}
M.~Singha and K.~Hemachandran, ``Content based image retrieval using color and
  texture,'' \emph{Signal \& Image Processing: An International Journal
  (SIPIJ)}, vol.~3, no.~1, pp. 39--57, 2012.

\bibitem{levkowitz1993glhs}
H.~Levkowitz and G.~T. Herman, ``Glhs: a generalized lightness, hue, and
  saturation color model,'' \emph{CVGIP: Graphical Models and Image
  Processing}, vol.~55, no.~4, pp. 271--285, 1993.

\bibitem{jain1997object}
A.~K. Jain, N.~K. Ratha, and S.~Lakshmanan, ``Object detection using gabor
  filters,'' \emph{Pattern Recognition}, vol.~30, no.~2, pp. 295--309, 1997.

\bibitem{palm2000gabor}
C.~Palm, D.~Keysers, T.~Lehmann, and K.~Spitzer, ``Gabor filtering of complex
  hue/saturation images for color texture classification,'' in \emph{Int. Conf.
  on Computer Vision}, vol.~2, 2000, pp. 45--49.

\bibitem{chen2008fast}
T.-W. Chen, Y.-L. Chen, and S.-Y. Chien, ``Fast image segmentation based on
  k-means clustering with histograms in hsv color space,'' in \emph{Multimedia
  Signal Processing, 2008 IEEE 10th Workshop on}.\hskip 1em plus 0.5em minus
  0.4em\relax IEEE, 2008, pp. 322--325.

\bibitem{bernardino2005real}
A.~Bernardino and J.~Santos-Victor, ``A real-time gabor primal sketch for
  visual attention,'' in \emph{Iberian Conference on Pattern Recognition and
  Image Analysis}.\hskip 1em plus 0.5em minus 0.4em\relax Springer, 2005, pp.
  335--342.

\bibitem{haghighat2013identification}
M.~Haghighat, S.~Zonouz, and M.~Abdel-Mottaleb, ``Identification using
  encrypted biometrics,'' in \emph{International Conference on Computer
  Analysis of Images and Patterns}.\hskip 1em plus 0.5em minus 0.4em\relax
  Springer, 2013, pp. 440--448.

\bibitem{fei2007learning}
L.~Fei-Fei, R.~Fergus, and P.~Perona, ``Learning generative visual models from
  few training examples: An incremental bayesian approach tested on 101 object
  categories,'' \emph{Computer Vision and Image Understanding}, vol. 106,
  no.~1, pp. 59--70, 2007.

\bibitem{krizhevsky2010convolutional}
A.~Krizhevsky and G.~Hinton, ``Convolutional deep belief networks on
  cifar-10,'' \emph{Unpublished manuscript}, vol.~40, 2010.

\bibitem{lecun1998gradient}
Y.~LeCun, L.~Bottou, Y.~Bengio, and P.~Haffner, ``Gradient-based learning
  applied to document recognition,'' \emph{Proceedings of the IEEE}, vol.~86,
  no.~11, pp. 2278--2324, 1998.

\bibitem{convnetJS}
A.~Karpathy, ``http://cs.stanford.edu/people/karpathy/convnetjs
  /demo/cifar10.html,'' 2015.

\bibitem{zagoruyko2016wide}
S.~Zagoruyko and N.~Komodakis, ``Wide residual networks,'' \emph{arXiv preprint
  arXiv:1605.07146}, 2016.

\bibitem{han2015learning}
S.~Han, J.~Pool, J.~Tran, and W.~Dally, ``Learning both weights and connections
  for efficient neural network,'' in \emph{Advances in Neural Information
  Processing Systems}, 2015, pp. 1135--1143.

\bibitem{collobert2002torch}
R.~Collobert, S.~Bengio, and J.~Mari{\'e}thoz, ``Torch: a modular machine
  learning software library,'' Idiap, Tech. Rep., 2002.

\bibitem{yuan2006automatic}
X.~Yuan, W.~Lai, T.~Mei, X.-S. Hua, X.-Q. Wu, and S.~Li, ``Automatic video
  genre categorization using hierarchical svm,'' in \emph{Image Processing,
  2006 IEEE International Conference on}.\hskip 1em plus 0.5em minus
  0.4em\relax IEEE, 2006, pp. 2905--2908.

\bibitem{gao2011discriminative}
T.~Gao and D.~Koller, ``Discriminative learning of relaxed hierarchy for
  large-scale visual recognition,'' in \emph{Computer Vision (ICCV), 2011 IEEE
  International Conference on}.\hskip 1em plus 0.5em minus 0.4em\relax IEEE,
  2011, pp. 2072--2079.

\end{thebibliography}
\end{document}